\documentclass[journal]{IEEEtran}

\usepackage{ifpdf}

\usepackage{cite}

\ifCLASSINFOpdf
  \usepackage[pdftex]{graphicx}
\else
  \usepackage[dvips]{graphicx}
\fi
\usepackage{amsmath}
\usepackage{algorithmic}
\usepackage{array}
\ifCLASSOPTIONcompsoc
  \usepackage[caption=false,font=normalsize,labelfont=sf,textfont=sf]{subfig}
\else
  \usepackage[caption=false,font=footnotesize]{subfig}
\fi
\usepackage{fixltx2e}
\usepackage{stfloats}
\ifCLASSOPTIONcaptionsoff
  \usepackage[nomarkers]{endfloat}
 \let\MYoriglatexcaption\caption
 \renewcommand{\caption}[2][\relax]{\MYoriglatexcaption[#2]{#2}}
\fi
\let\MYorigsubfloat\subfloat
\renewcommand{\subfloat}[2][\relax]{\MYorigsubfloat[]{#2}}
\usepackage{url}
\usepackage{booktabs}
\usepackage{multirow}

\begin{document}
%
% paper title Multi-Modal Sensor Fusion Enabled Deep Neural Networks for End-to-end Autonomous Driving with Scene Understanding
\title{Multi-modal Sensor Fusion-Based Deep Neural Network for End-to-end Autonomous Driving with Scene Understanding}

% a structure at a ~ so this keeps an author's name from being broken across two lines.

\author{Zhiyu Huang, %~\IEEEmembership{Member,~IEEE,}
        Chen Lv,~\IEEEmembership{Senior Member,~IEEE}, Yang Xing, Jingda Wu%
        %and~Jane~Doe,~\IEEEmembership{Life~Fellow,~IEEE}% <-this % stops a space
\thanks{Z. Huang, C. Lv, Y. Xing and J. Wu are with the School of Mechanical and Aerospace Engineering, Nanyang Technological University, Singapore, 639798. (e-mail: \{zhiyu001, jingda001\}@e.ntu.edu.sg, \{lyuchen, yang.xing\}@ntu.edu.sg)}%
\thanks{This work was supported by the SUG-NAP Grant (No. M4082268.050) of Nanyang Technological University, Singapore.}%
\thanks{Corresponding author: C. Lv}

}
\maketitle

\begin{abstract}
This study aims to improve the performance and generalization capability of end-to-end autonomous driving with scene understanding leveraging deep learning and multimodal sensor fusion techniques. The designed end-to-end deep neural network takes as input the visual image and associated depth information in an early fusion level and outputs the pixel-wise semantic segmentation as scene understanding and vehicle control commands concurrently. The end-to-end deep learning-based autonomous driving model is tested in high-fidelity simulated urban driving conditions and compared with the benchmark of CoRL2017 and NoCrash. The testing results show that the proposed approach is of better performance and generalization ability, achieving a 100\% success rate in static navigation tasks in both training and unobserved situations, as well as better success rates in other tasks than the prior models. A further ablation study shows that the model with the removal of multimodal sensor fusion or scene understanding pales in the new environment because of the false perception. The results verify that the performance of our model is improved by the synergy of multimodal sensor fusion with scene understanding subtask, demonstrating the feasibility and effectiveness of the developed deep neural network with multimodal sensor fusion.
\end{abstract}

% Note that keywords are not normally used for peer review papers.

% \fi
\IEEEpeerreviewmaketitle

\section{Introduction}
\label{sec1}

\IEEEPARstart{C}{urrently}, the solutions to autonomous driving can be divided into two streams: modular pipeline and end-to-end sensorimotor. The modular solution divides the overall driving task into multiple subtasks, including perception, localization, decision making, path planning, and motion control \cite{schwarting2018planning}. Thanks to the rapid development of machine learning, the end-to-end solution can omit all the subtasks of the modular solution by building a mapping from high dimensional raw inputs of sensors directly to vehicle control command outputs \cite{bojarski2016end}. In this way, the end-to-end solution can significantly reduce the complexity of the autonomous driving system while realizing the feasibility and effectiveness of the task objective. Recent end-to-end autonomous driving arises with the prosperity of deep learning techniques. It starts with a project from NVIDIA \cite{bojarski2016end}, in which the researchers used a convolutional neural network (CNN) that only takes as input the monocular image of the frontal road to steer a commercial vehicle. The end-to-end driving system has shown its ability in various driving conditions, including highways and rural roads. Inspired by this work, many works adopt the end-to-end approach to tackle autonomous driving problems, from lane-keeping \cite{chen2017end} to driving in rural roads \cite{kendall2019learning} and complexed urban areas \cite{hawke2019urban}.

End-to-end autonomous driving utilizes a deep neural network, which takes raw data from sensors (e.g., images from cameras, point cloud from Lidar) as inputs and outputs control commands (e.g. throttle, brake, and steering angle), to drive a vehicle. Although it does not explicitly divide the driving task into several sub-modules, the network itself can be divided into two coupled parts in terms of functionalities, i.e. the environmental perception part and the driving policy part. The perception part is usually a CNN, which takes an image as the input and generates a low dimensional feature representation. The latent feature representation is subsequently connected to the driving policy module, which is typically a fully connected network, generating control commands as output. The vehicle control commands are used as the only supervision signals in a large amount of literature. These control signals are capable of supervising the network to learn driving commands while finding features related to driving tasks \cite{yang2017feature}, such as lane markers and obstacles \cite{bojarski2017explaining}. However, such supervisions may not be strong enough to yield a good latent representation of the driving scene, and thus result in overfitting and lack of generalization capability and deteriorate the distributional shift problem \cite{codevilla2019exploring}. Therefore, learning a good latent representation of the driving scene is of great importance for further improving the performance of end-to-end autonomous driving.

So as to learn a good representation of the driving environment, several factors that are learned from the naturalistic human driving need to be considered. First, humans perceive the environment with stereo vision, which brings in the depth information. That means humans observe the driving scene with multimodal information. From the perspective of neural networks, multimodal information can bring diverse and vibrant information of the driving environment to the latent representation and consequently lead to a better driving policy. Therefore, the multimodal sensor fusion technique is necessary to fuse vision and depth information for end-to-end autonomous driving. In \cite{xiao2019multimodal}, the authors fuse RGB image and corresponding depth map into the network to drive a vehicle in a simulated urban area and thoroughly investigate different multimodal sensor fusion methods, namely the early, mid, and late fusion and their influences on driving tasks. The end-to-end network designed in \cite{amado2019end} works with RGB and depth images measured from a Kinect sensor to steer a $1/4$ scale vehicle in the real world. Sobh et al. \cite{sobh2018end} proposes a deep neural network with late fusion design that ingests semantic segmentation maps and different representations of LiDAR point clouds to generate driving actions. Chen et al. \cite{chen2018lidar} also use camera images and point clouds as the inputs of their end-to-end driving network, while they use PointNet to directly process disordered point clouds. The results from these studies conclusively show that fusing camera images and depth information is more advantageous over the method using RGB image as a single-modal input in terms of the driving performance.

However, multimodal sensor fusion cannot guarantee a performance improvement if only the control commands act as the supervision signals. This is because of the causal confusion \cite{codevilla2019exploring}, which means the network might make spurious correlations between some specific feature patterns and the control commands. With more modalities, the network may only rely on some of the features from particular modalities to make decisions and ignore the information from other modalities. To solve this problem, we can think of another factor learned from human driving, which is to understand the driving scene and then make decisions. For neural networks, this can be achieved by training the end-to-end driving network with auxiliary tasks, to explicitly shape the latent representation by incorporating key features of the driving scene. For example, Xu et al. \cite{xu2017end} adds a semantic segmentation loss as a regularization when training an end-to-end network to predict the steering angle using visual images. In \cite{li2018rethinking}, the authors first pre-train an encoder-decoder structure, which takes a monocular camera image as the input and outputs the semantic segmentation and depth estimation of the image. The encoded latent representation from the penultimate layer of the encoder then inputs to the driving policy network. Leveraging a similar encoder-decoder structure, the network designed in \cite{hawke2019urban} receives an image as observation and performs semantic segmentation, depth prediction, and optical flow estimation first, and then uses the latent representation from the encoder to train a policy network to drive a real car in urban streets. Wang et al. \cite{wang2019deep} proposes a method to train an end-to-end driving network with object detection task, and both the object-centric features and latent contextual features are used to control a vehicle. These studies suggest that introducing the basic knowledge of driving scenes into the network would result in better generalization ability and robustness in an unobserved environment. The main limitation of these works, however, is that they only take one single modality (visual image) into account and thus does not fully utilize the capability of the multimodal information.

To solve the above issues and further improve the performance, this study develops a deep neural network with multimodal sensor fusion for end-to-end autonomous driving with scene understanding. The main contributions of this paper are listed as follows.

\begin{enumerate}
\item A deep neural network is proposed for end-to-end autonomous driving with scene understanding, which consists of three coupled parts in functionality. The common component is the multimodal sensor fusion encoder, which fuses multi-source sensor data, including the visual image and depth map, at an early stage to form low-dimensional latent features. The scene understanding is realized by performing pixel-wise semantic segmentation on the image following a convolutional encoder-decoder structure \cite{badrinarayanan2017segnet}, while the driving policy follows the conditional imitation learning paradigm\cite{codevilla2018end}.

\item The end-to-end driving system controlled by proposed network is tested in high-fidelity simulated urban driving scenarios with the CoRL2017 and NoCrash benchmarks.

\item An ablation test is conducted to investigate the effects of multimodal sensor fusion and scene understanding auxiliary task on the end-to-end driving performance.
\end{enumerate}

The remainder of this paper is organized as follows. Section \ref{sec2} introduces the method of multimodal sensor fusion with scene understanding and depicts the imitation learning framework for driving policy. Section \ref{sec3} details the data collection process, the training protocol of the neural network and the testing protocol. Section \ref{sec4} describes the analysis of the experiment results and ablation test. Finally, conclusions are summarized in Section \ref{sec5}.

\section{Methodology}
\label{sec2}

\begin{figure}[htp]
    \centering
    \includegraphics[width=\linewidth]{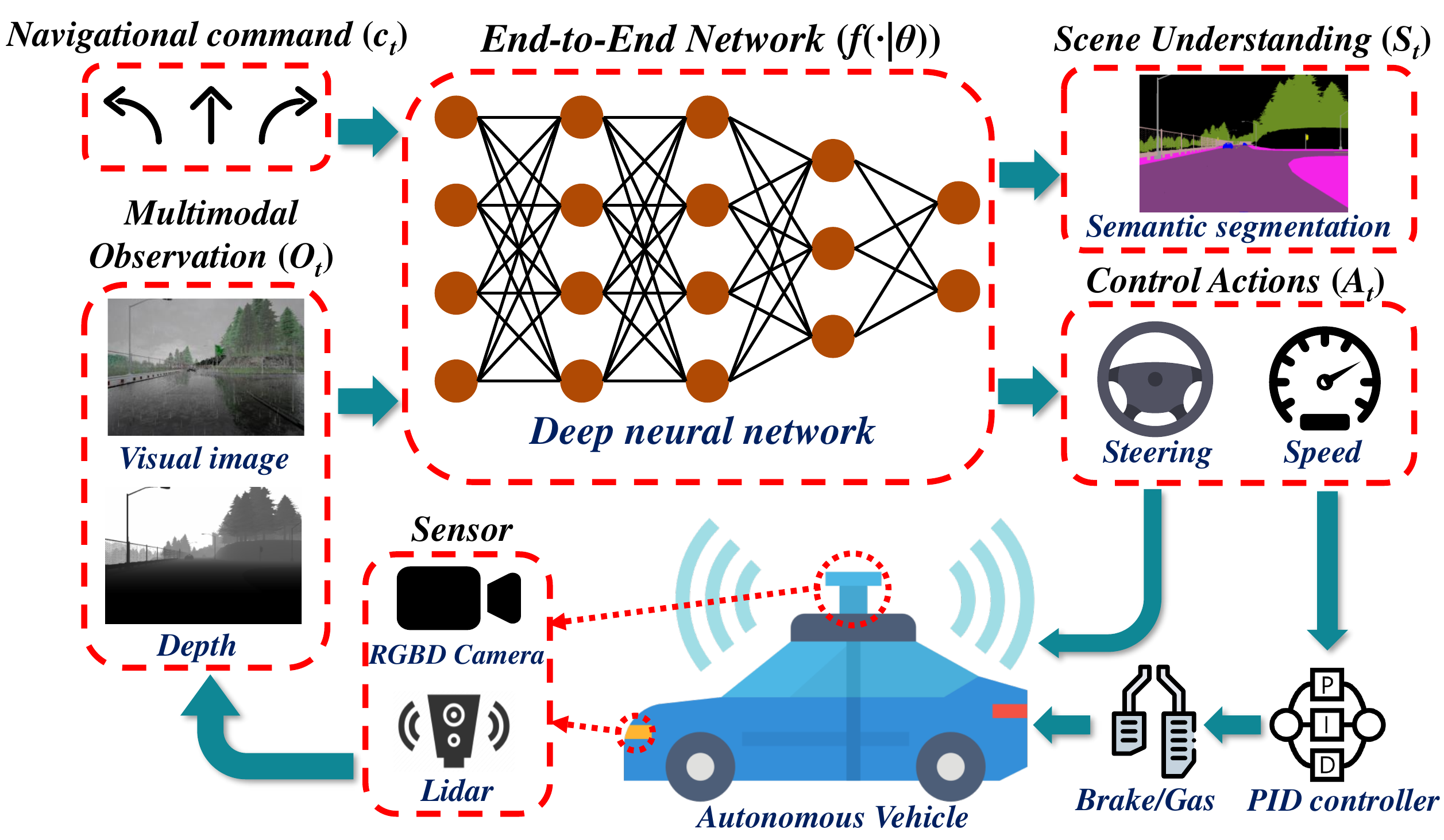}
    \caption{The framework of our multimodal sensor fusion with scene understanding end-to-end driving model.}
    \label{fig1}
\end{figure}

The end-to-end autonomous driving with scene understanding task is modeled as a mapping parameterized by $\theta$ from multimodal observation $O_t$, the navigational guidance $c_t$ to driving command $A_t$ and the scene understanding $S_t$ at timestep $t$, i.e., ${A_t},{S_t} = f(\left. {{O_t}}, c_t \right|\theta )$. The framework of the developed end-to-end driving model is shown in Fig. \ref{fig1}. The end-to-end deep neural network consists of two parts, which are the multimodal sensor fusion with scene understanding and the driving policy. The end-to-end deep neural network takes the multimodal sensor data and navigational direction as inputs and generates the semantic segmentation map and steering and speed control commands as outputs. The speed control is then converted into the throttle and brake control signals via a low-level PID controller. We will describe the details as follows. 

\subsection{Multimodal sensor fusion with scene understanding}
Let the raw sensor inputs be ${O_t} = \left\{ {o_t^i} \right\}_{i = 1}^N$, which consists of $N$ various observation modalities ${o_t^i}$ measured from different sensors with multiple modalities, such as camera, Lidar, or odometer. These modalities, for example, could be RGB image, depth map, point clouds, and vehicle speed. We use a multimodal sensor fusion encoder network with parameter $\theta_E$, denoted as $f\left( {\left. \cdot  \right|{\theta _E}} \right)$, to encode the high dimensional raw inputs from the sensors $O_t$ into a low dimensional latent representation $z_t$. This sensor fusion process is formulated as

\begin{equation}
\label{eq1}
{z_t} = f(\left. {{O_t}} \right|{\theta _E}).   
\end{equation}

To incorporate the scene understanding in the overall end-to-end driving network, we introduce a decoder network $f\left( {\left. \cdot  \right|{\theta _D}} \right)$ parameterized by $\theta_D$. The decoder projects the latent representation $z_t$ to a high dimensional representation denoted as ${S_t} = \left\{ {s_t^i} \right\}$, in which $s_t^i$ conveys an understanding of the driving scene in various formats. Such a high-dimension representation could be the reconstructed image, pixel-wise image semantic segmentation, or point-wise point cloud segmentation. This scene understanding process is shown in the following equation:

\begin{equation}
{S_t} = f\left( {\left. {{z_t}} \right|{\theta _D}} \right).    
\end{equation}

In this way, the latent representation $z_t$ is explicitly shaped to incorporate the driving scene understanding, and it is subsequently used for the downstream driving task described in the next subsection.

\subsection{Conditional driving policy}
To resolve the ambiguity of the route selection in an intersection, the navigational commands can be incorporated into the network \cite{codevilla2018end}. Let the high-level navigational command be $c_t$, which represents the different directional guidance, such as “go straight” and “turn left”. The driving policy is modeled as a neural network $f\left( {\left. \cdot  \right|{\theta _P}} \right)$ with parameter $\theta_P$. Given the latent representation $z_t$ of the driving scene, the driving policy takes $z_t$ and navigational command $c_t$ as inputs and generates the low-level control commands denoted by $A_t$, as the output to the vehicle. The control commands in this paper are the steering angle and the longitudinal speed, which are denoted as $a_t^{{\text{steer}}}$ and $a_t^{{\text{speed}}}$, respectively, and thus ${A_t} = \left\{ {a_t^{{\text{steer}}},a_t^{{\text{speed}}}} \right\}$. The model of the conditional driving policy is given by:

\begin{equation}
{A_t} = f \left( {\left. {{z_t},{c_t}} \right|{\theta _P}} \right).
\end{equation}

\subsection{End-to-end learning}
Although the scene understanding network with sensor fusion and the driving policy network can be trained sequentially, joint training is more efficient and can obtain task-specific features. Therefore, we integrate the different parts, namely the multimodal sensor fusion encoder, scene understanding decoder, and the conditional driving policy, as a whole and train the entire network in an end-to-end manner.

Let the overall network be denoted as $f\left( {\left. \cdot  \right|{\theta }} \right)$ with the parameter $\theta {\text{ = }}\left\{ {{\theta _E},{\theta _D},{\theta _P}} \right\}$. Thus the input of this network becomes $\left\{ {{O_t},{c_t}} \right\}$, and thereby the end-to-end driving network can be expressed by:

\begin{equation}
{S_t},{A_t} = f\left( {\left. {{O_t}, c_t} \right|\theta } \right).
\end{equation}

\begin{figure*}[htp]
    \centering
    \includegraphics[width=0.9\linewidth]{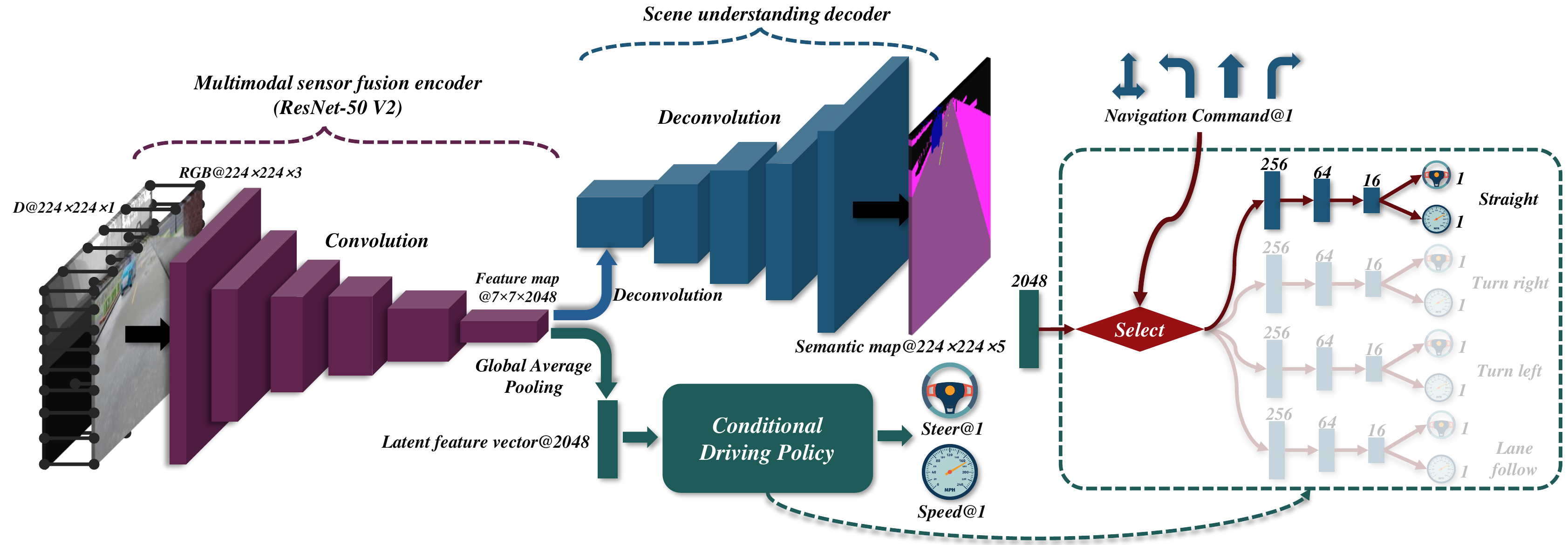}
    \caption{The structure of the deep neural network with multimodal sensor fusion for end-to-end autonomous driving with scene understanding.}
    \label{fig2}
\end{figure*}

Provided the dataset $\mathcal{D} = \left\{ {{O_t},{A_t},{S_t},{c_t}} \right\}_{t = 1}^L$ of the multimodal observations $O_t$ and the associated scene understanding representation $S_t$ such as semantic maps, as well as recorded driving data, i.e., the steering and speed control actions $A_t$ and navigational guidance $c_t$, the network is trained to understand the scene and learn driving from demonstrations simultaneously. The learning process is to optimize the network parameter $\theta$ with respect to the following function:

\begin{equation}
\label{eq5}
\mathop {\min }\limits_\theta  {\text{ }}{\lambda _1}{\mathcal{L}_{steer}} + {\lambda _2}{\mathcal{L}_{speed}} + {\lambda _3}{\mathcal{L}_{scene}},
\end{equation}
where ${\mathcal{L}_{steer}}$  and ${\mathcal{L}_{speed}}$ are loss functions for steering and speed control actions, and ${\mathcal{L}_{scene}}$ is the loss function for the scene understanding task;  $\lambda_1$, $\lambda_2$ and $\lambda_3$ are the weights for individual output loss to balance the scale and emphasize the important loss function. The individual loss function is listed as follows:

\begin{equation}
\label{eq6}
{\mathcal{L}_{steer}} = \frac{1}{L}\sum\limits_{i = 1}^L {{{\left( {1 + \alpha {{\left| {a_i^{{\text{steer,gt}}}} \right|}^\beta }} \right)}^\gamma }} {\left( {a_i^{{\text{steer}}} - a_i^{{\text{steer,gt}}}} \right)^2},    
\end{equation}

\begin{equation}
\label{eq7}
{\mathcal{L}_{speed}} = \frac{1}{L}\sum\limits_{i = 1}^L {{{\left( {a_i^{{\text{speed}}} - a_i^{s{\text{peed,gt}}}} \right)}^2}}, 
\end{equation}

\begin{equation}
\label{eq8}
{\mathcal{L}_{scene}}{\text{ = }}\frac{1}{L}\frac{1}{{{N_p}}}\frac{1}{{{N_c}}}\sum\limits_{i = 1}^L {\sum\limits_{j = 1}^{{N_p}} {\sum\limits_{k = 1}^{{N_c}} { - s_{ijk}^{gt}\log {s_{ijk}}} } },  
\end{equation}
where $L$ is the total number of the training samples, and the notations with $gt$ indicate the ground-truth values. After some initial trials, the loss used in \cite{yuan2019steeringloss} (Eq. \ref{eq6}) and the L2 loss are adopted, respectively, for the steering and speed control actions in this study (Eq. \ref{eq7}). Since most of the training samples are with small steering values, using Eq. \ref{eq6} can enlarge the impact of sharp steering value, which helps to train the network. Here we choose $\alpha  = 5$, $\beta=1$, and $\gamma=2$ for Eq. \ref{eq6}. For scene understanding learning, we use pixel-wise cross-entropy loss (Eq. \ref{eq8}) for the image semantic segmentation task. In Eq. \ref{eq8}, $s_{ijk}$ stands for the probability of a pixel $j$ of an image $i$ which belongs to category $k$, and $N_c$ is the number of different categories, and $N_p$ is the number of pixels in an image.

\subsection{The neural network architecture}
In this paper, the input multimodal sensor data is the RGB image and its associated depth map recording the driving scene ahead of the ego vehicle. The depth map encodes the depth information of each pixel in the RGB image. Therefore, the observation is ${O_t} = \left\{ I_t, D_t \right\}$ and the scene understanding representation $S_t$ is the semantic map, which shows the category of each pixel in the input image.

Fig. \ref{fig2} shows the structure of the proposed deep neural network with multimodal sensor fusion for end-to-end autonomous driving and scene understanding. It can be implicitly divided into three parts: multimodal sensor fusion encoder, scene understanding decoder, and conditional driving policy. RGB image $I^{W \times H \times 3}$ and the depth map $D^{W \times H \times 1}$ with width $W$ and height $H$ ($224\times224$ in our case) are concatenated channel-wise first to form an RGBD structure, and then the RGBD data is input to the multimodal sensor fusion encoder which follows a ResNet-50 V2 structure \cite{he2016identity}. The output of the encoder is a feature map, which is then connected to a scene understanding decoder that consists of five deconvolutional layers with softmax activation function for the last deconvolutional layer. Each deconvolutional layer except the last one is followed by a batch normalization layer and then the ReLU nonlinearity. The numbers of filters of the five deconvolutional layers are 512, 128, 64, 16, 5, respectively. The kernel size of the filters in each deconvolutional layer is set as 3, and the strides are 4, 2, 2, 2, 1, respectively. Eventually, the decoder outputs the category of each pixel in the original image to express its understanding of the driving scene. Five categories, namely the lane, road line, sidewalk, vehicles or pedestrians, and others, are chosen. The feature map is global average pooled to generate a latent feature vector, followed by the conditional driving policy network to compute driving commands concurrently.

The conditional driving policy is a branched fully connected network. The input of the driving policy is the latent feature vector derived from the multimodal sensor fusion encoder. The navigational command $c_t$ activates the corresponding branch, and each dedicated branch is a fully connected network that outputs the desired speed and steering control signals $A_t$ for specific guidance. In other words, the navigational command can be seen as a switch that selects which branch outputting the control commands. We have four navigation commands, which are “straight”, “lane follow”, “turn right” and “turn left”. Each dense layer uses the ReLU activation function and a dropout rate of 0.5 except the last dense layer that generates the control commands.

\section{Model Training and Testing}
\label{sec3}

\begin{figure*}[htp]
    \centering
    
    \subfloat[]{\includegraphics[width=0.45\linewidth]{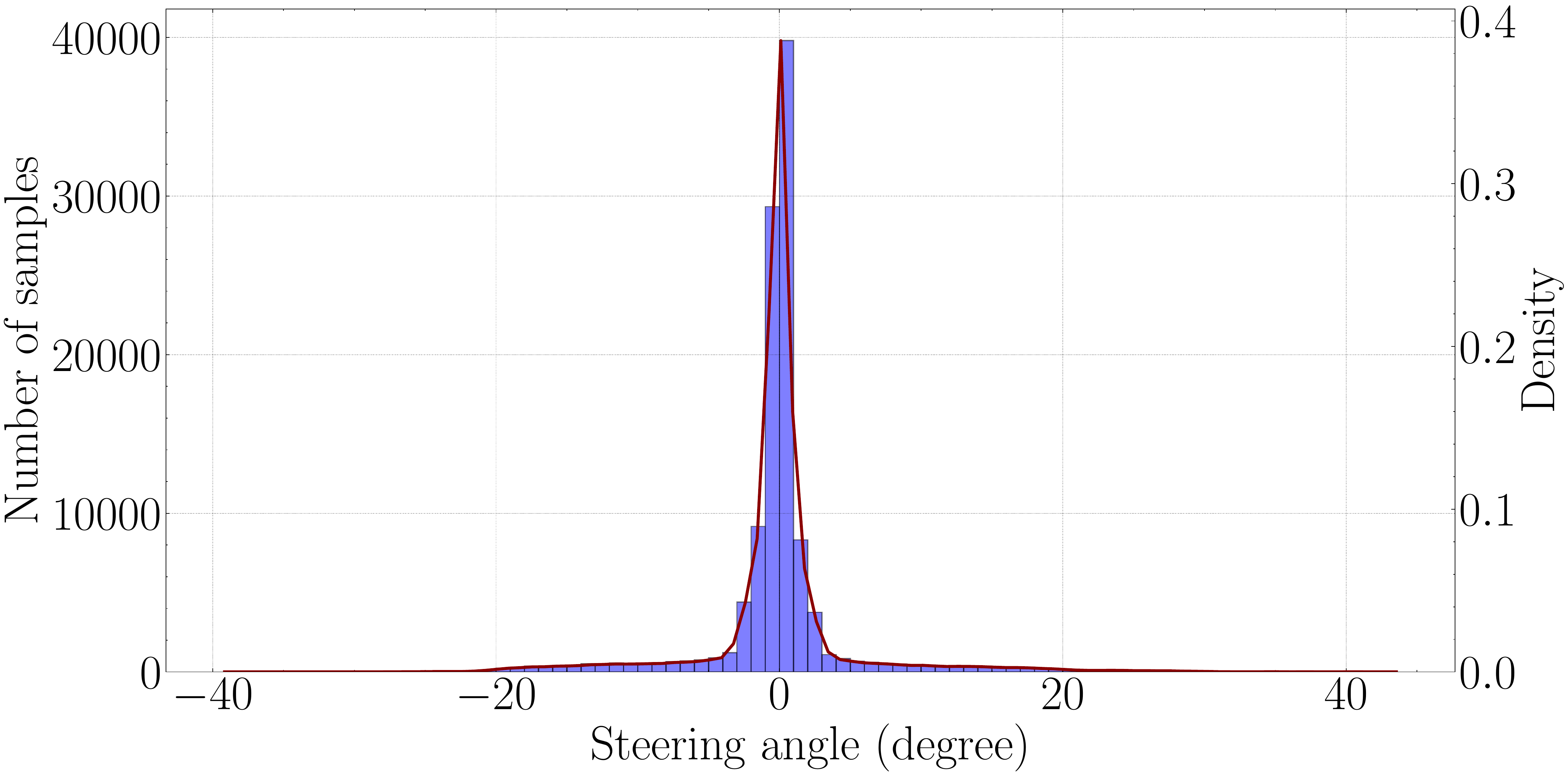}%
    \label{fig3a}}
    \hfil
    \subfloat[]{\includegraphics[width=0.45\linewidth]{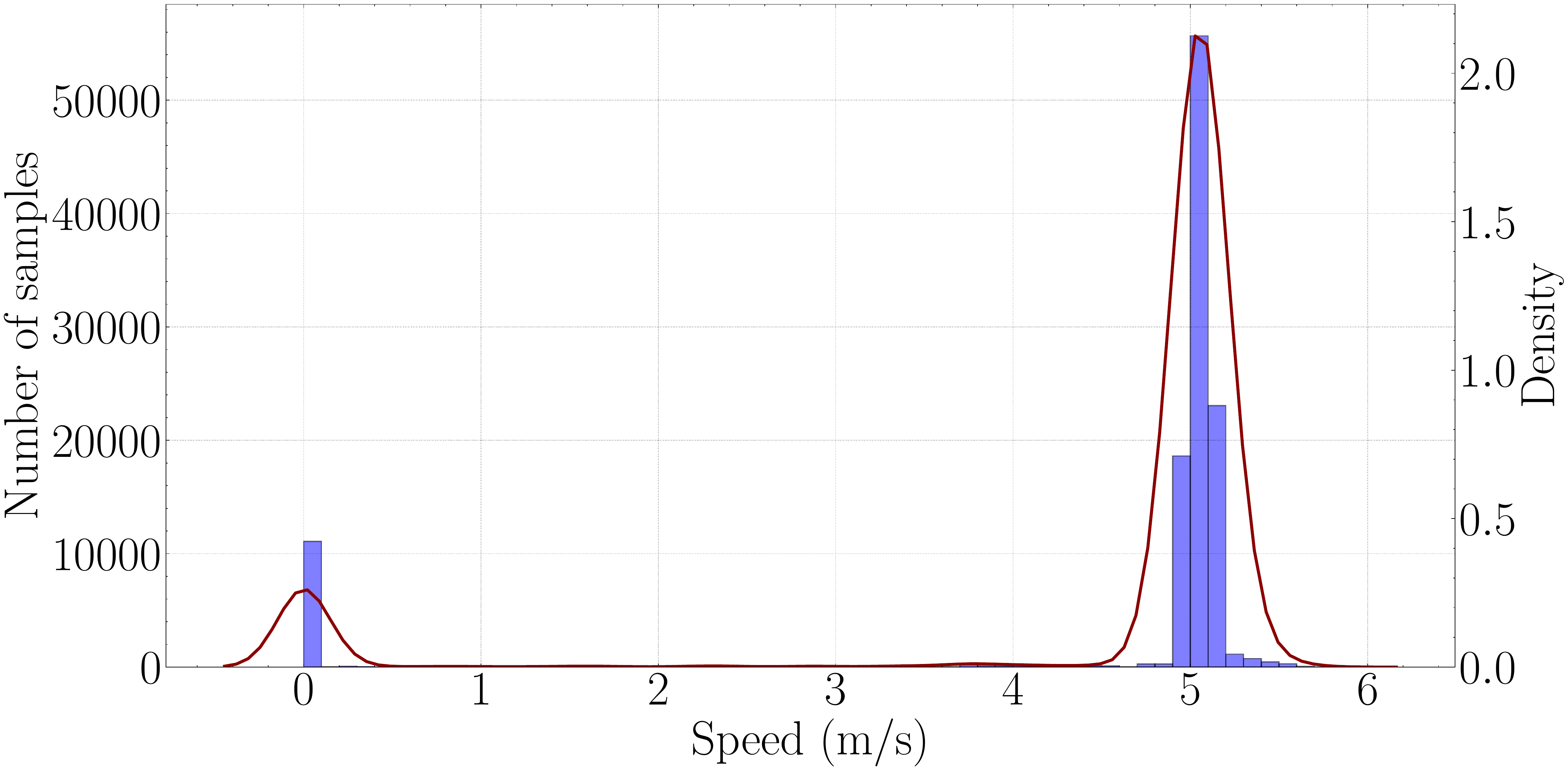}%
    \label{fig3b}}
    
    \subfloat[]{\includegraphics[width=0.45\linewidth]{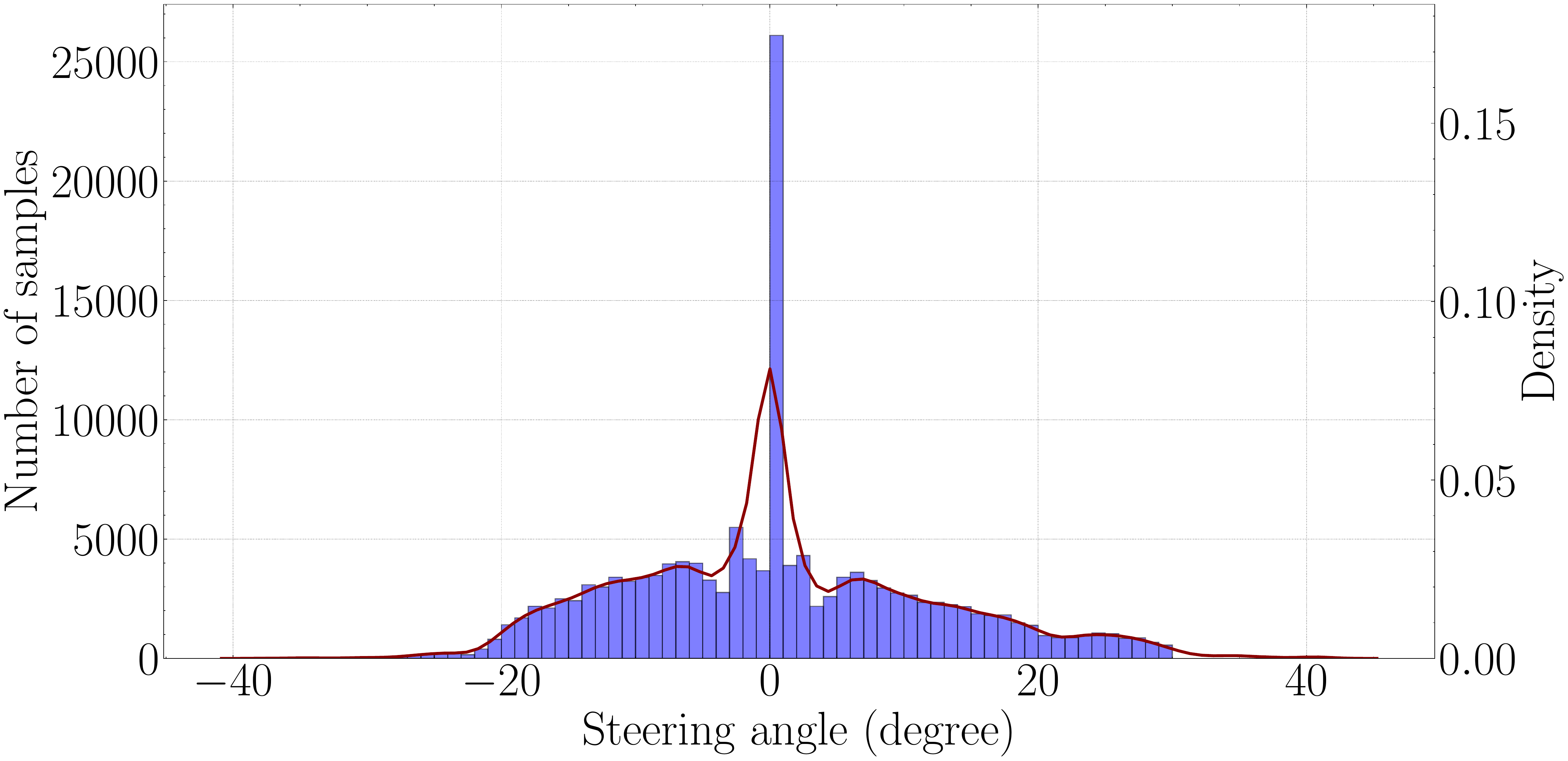}%
    \label{fig3c}}
    \hfil
    \subfloat[]{\includegraphics[width=0.45\linewidth]{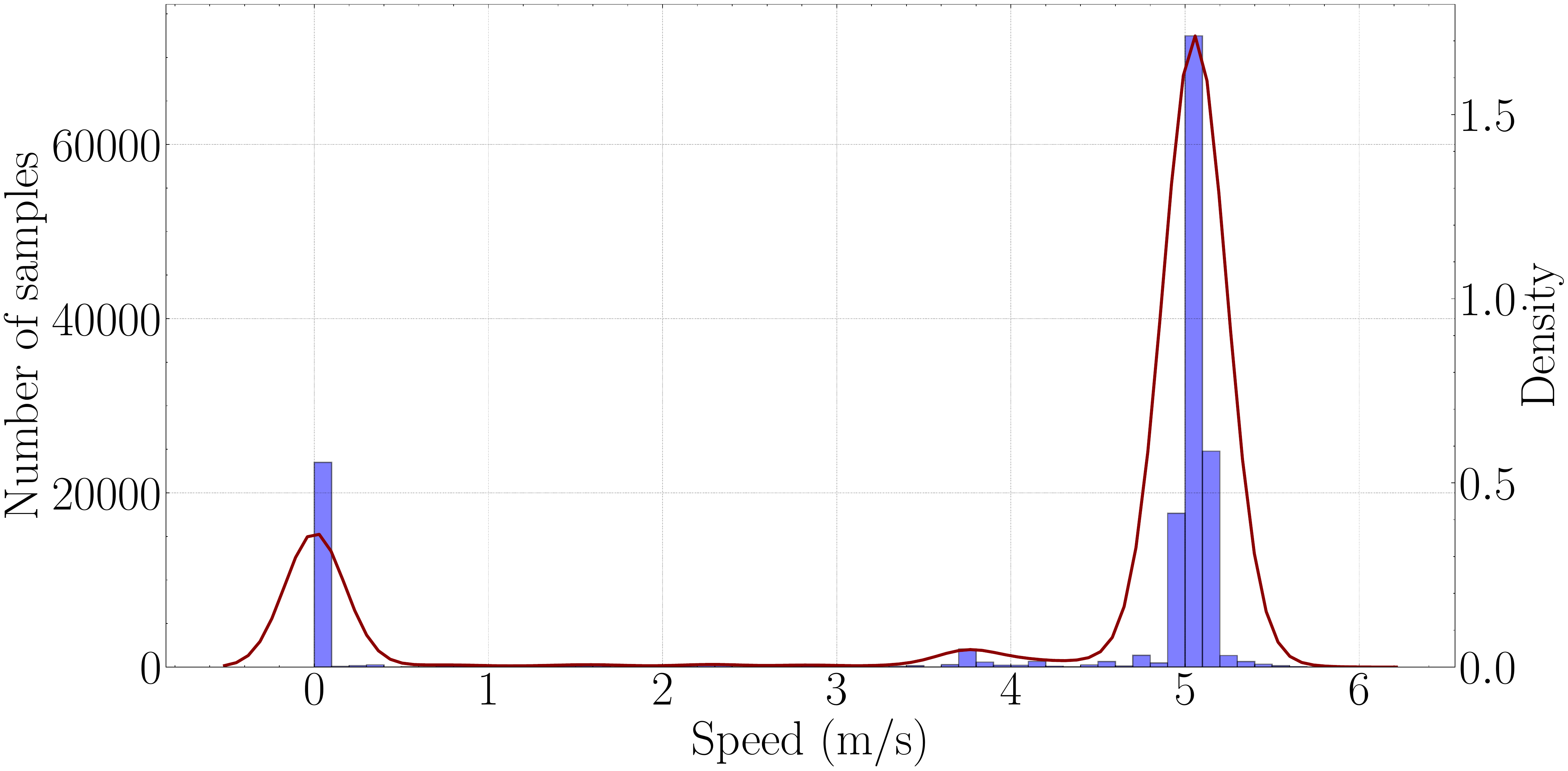}%
    \label{fig3d}}

    \caption{Dataset balancing for lane-keeping: (a) Original steering angle distribution; (b) Original speed distribution; (c) Processed steering angle distribution; (d) Processed speed distribution.}
    \label{fig3}
\end{figure*}

\subsection{Data collection}
The dataset is collected in two urban driving scenarios, i.e. the Town 01 and Town 02, in CARLA simulator \cite{dosovitskiy2017carla}. They are two different small urban areas with two-lane roads, curves, and intersections, and the weather conditions in the scenarios can be dynamically adjusted. These two towns have different road layouts. Town 01 is larger with 2.9 km of road and 11 intersections, while Town 2 deploys 1.4 km of road and eight intersections. The visual nuisances in the towns, such as static obstacles and buildings, are also different from each other. We first spawn 40 vehicles and 40 pedestrians in the town and use the built-in autopilot function to control one of the vehicles to run from one target position to another. A depth-sensing camera is mounted at the front end of the vehicle, capturing RGB images of the frontal road with a resolution of $800 \times 600$ at 10 FPS, as well as the corresponding depth maps and ground truth semantic maps. The measurements of the vehicle states, including the navigational direction from a route planner, vehicle speed, heading, position, and steering command, are recorded and synchronized with the sensor data. We inject a one-second random noise into the steering control every 5 seconds to purposely make the vehicle deviate its normal driving path and thereby enrich the dataset with samples of correcting errors. This will help overcome the distributional shift problem in imitation learning. The samples with noise injection from the collected data need to be eliminated to construct the training dataset. Overall, the total data volume collected in Town 01 is 132233, corresponding to approximately 3.7 hours of driving, and 28716 collected for Town 02. Only the data collected from Town 01 is used as the training data, while the data obtained from Town 02 is used for validation. Therefore, in the rest of this paper, Town 01 is denoted as the training scenario and Town 02 is denoted as the testing scenario.

\subsection{Model training}
Balancing the training data is an essential step before training since the vehicle goes straight at most of the time, and consequently, small values dominate the distribution of the steering angles, as shown in Fig. \ref{fig3a}. Note that we display the actual steering angle of the vehicle in degrees by multiplying 70 to the normalized steering angle originally provided by the simulator. Since the majority of the dataset is filled with lane-keeping cases, we focus on balancing the distribution of steering and speed controls concerning the lane-keeping command. As presented in Fig. \ref{fig3a}, the steering angle between -5 and 5 degrees account for nearly 90\% of the data. Therefore, we need to downsample the majority class (range) and upsample the minority class (range). We randomly discard 80\% of the data points in which steering angle falls within  -5 and 5 degrees and duplicate the rest data points six times and then append them in the training dataset. However, this process brings an imbalance in speed distribution, i.e., significantly increasing the number of data points with high speed. This may lead to a problem that the agent cannot learn to stop. Thus, we duplicate the processed data points with speed being below 1 m/s three times. This will increase the data points with small steering angles, thereby inducing another imbalance. However, it is a trade-off that should be addressed. Eventually, the processed data distribution for lane-keeping is shown in Fig. \ref{fig3c} and Fig. \ref{fig3d}. 

We choose the weights for the loss function (Eq. \ref{eq5}) as $\left\{ {{\lambda _{\text{1}}}{\text{ = 10, }}{\lambda _2}{\text{ = 1, }}{\lambda _3}{\text{ = 2}}} \right\}$ after careful trails. We crop top half of the collected RGB images and depth images, and resize them to $224 \times 224$ and normalize the channel values to $[0,1]$. During training, we apply Gaussian noise, coarse dropout, contrast normalization, Gaussian blur on the RGB images with a probability of 0.1 for data augmentation.

The neural network is trained with NAdam optimizer with an initial learning rate of 0.0003. The learning rate is scaled by a factor of 0.5 of the previous learning rate, if the validation loss does not drop down for consecutive five epochs. The batch size is set to 32, and the total training epoch is set to 100. Early stopping is used, i.e., terminating training if the validation loss does not improve for 20 epochs, to avoid overfitting. The network is trained on an NVIDIA RTX 2080Ti GPU. The model with the least validation loss is saved and used for model testing. 

\subsection{Model testing}
The model testing is carried out in both the training scenario and the testing scenario with different weather conditions. We utilize the end-to-end driving model to control the vehicle, navigating it from the predefined starting point to the target ending point. At each timestep, the end-to-end agent receives multimodal sensor inputs (i.e. the RGB image and depth map) and a high-level navigational command to compute the desired speed and steering control signals, as well as the semantic segmentation result to show its understanding of the driving scene. To convert the speed control action to vehicle control commands, we use a PID controller in the low-level to compute throttle and brake controls for tracking the desired speed.

The evaluation of the proposed end-to-end autonomous driving follows the paradigms of the CoRL2017 benchmark \cite{dosovitskiy2017carla} and NoCrash benchmark \cite{codevilla2019exploring}. In the CoRL2017 benchmark, the agent should fulfill four driving tasks, and each task consists of distinct 25 predefined navigation routes. The four driving tasks are driving straight, turning, navigating in empty traffic, and navigating with dynamic traffic participants. A trail is considered a failure if the agent is stuck somewhere and exceeds the allowed time to reach the target point. The stricter and harder NoCrash benchmark consists of three driving tasks, and each task shares the same 25 predefined routes but differs in the densities of traffic, which are no traffic, regular traffic, and dense traffic, respectively. A trial is seen as fail if any collision happens or exceeding the time limit. 

As for the weather settings, the training scenario is with four weather settings that have been used for the collection of the training data. The four weather conditions are “clear afternoon”, “wet road afternoon”, “hard rain afternoon”, and “clear sunset”. The testing scenario has two different weather settings. For the CoRL2017 benchmark, the weather settings are “cloudy afternoon with the wet road” and “soft rain sunset”, while for the NoCrash benchmark, the weather settings include “wet road sunset” and “soft rain sunset”.

\section{Testing Results and Discussions}
\label{sec4}

\subsection{Testing results}
The main evaluation metric of the end-to-end autonomous driving model is selected as the success rate, which is defined as the percentage of episodes that the end-to-end driving agent has successfully completed in a task. Table \ref{table1} shows a comparison of our method with other state-of-the-art methods on the CoRL2017 benchmark. We select six other works listed in Table \ref{table1} for comparison, which are controllable imitative reinforcement learning (CIRL) \cite{liang2018cirl}, conditional affordance learning (CAL) \cite{sauer2018conditional} multi-task learning (MT) \cite{li2018rethinking}, conditional imitation learning with ResNet architecture and speed prediction (CILRS) \cite{codevilla2019exploring}, and multimodal early fusion (MEF) \cite{xiao2019multimodal}. The proposed method is denoted as MSFSU, which stands for multimodal sensor fusion with scene understanding. Table \ref{table2} shows the results of the success rate with different methods on the NoCrash benchmark.
To validate and compare the model performance in various conditions, we define multiple navigation tasks. The defined static navigation tasks consist of straight, one turn and navigation in the CoRL2017 benchmark and navigation in empty traffic in the NoCrash benchmark, while the dynamic navigation tasks are defined as containing navigating with dynamic traffic participants in the CoRL2017 benchmark and navigating in regular and densed traffic in the NoCrash benchmark.

\begin{table}[htp]
\centering
\caption{Comparison of success rate (\%) with different models on the CoRL2017 benchmark}
\label{table1}
\scalebox{0.85}{
\begin{tabular}{@{}ccccccccc@{}}
\toprule
Task                             & Scenario                                                           & CIRL   & CAL          & MT   & CILRS   & MEF  & MSFSU       \\ \midrule
Straight                         & \multirow{4}{*}{\begin{tabular}[c]{@{}c@{}}Training \\scenario \end{tabular}} & 98     & \textbf{100} & 96   & 96      & 99   & \textbf{100} \\
One turn                         &                                                                    & 97     & 97           & 87   & 92      & 99   & \textbf{100} \\
Navigation                       &                                                                    & 93     & 92           & 81   & 95      & 92   & \textbf{100} \\
Nav. dynamic                     &                                                                    & 82     & 83           & 81   & 92      & 89   & \textbf{98}  \\ \midrule
\multicolumn{1}{c}{Straight}     & \multirow{4}{*}{\begin{tabular}[c]{@{}c@{}}Testing \\scenario \end{tabular}}  & 98     & 94           & 96   & 96      & 96   & \textbf{100} \\
\multicolumn{1}{c}{One turn}     &                                                                    & 80     & 72           & 82   & 92      & 84   & \textbf{100} \\
\multicolumn{1}{c}{Navigation}   &                                                                    & 68     & 68           & 78   & 92      & 90   & \textbf{100} \\
\multicolumn{1}{c}{Nav. dynamic} &                                                                    & 62     & 64           & 62   & 90      & \textbf{94} & \textbf{94}  \\ \bottomrule
\end{tabular}}
\end{table}

\begin{table}[htp]
\caption{Comparison of success rate (\%) with different models on the NoCrash benchmark}
\label{table2}
\centering
\begin{tabular}{@{}ccccccc@{}}
\toprule
Task    & Scenario                                                           & CAL      & MT       & CILRS     & MSFSU              \\ \midrule
Empty   & \multirow{3}{*}{\begin{tabular}[c]{@{}c@{}}Training\\ scenario \end{tabular}} & $81\pm1$ & $84\pm1$ & $97\pm2$  & $\textbf{100}\pm0$ \\
Regular &                                                                    & $73\pm1$ & $54\pm1$ & $83\pm0$  & $\textbf{91}\pm1$  \\
Dense   &                                                                    & $42\pm1$ & $13\pm1$ & $42\pm2$  & $\textbf{61}\pm3$  \\ \midrule
Empty   & \multirow{3}{*}{\begin{tabular}[c]{@{}c@{}}Testing\\ scenario\end{tabular}}   & $25\pm3$ & $57\pm0$ & $90\pm2$  & $\textbf{100}\pm0$ \\
Regular &                                                                    & $14\pm2$ & $32\pm2$ & $56\pm2$  & $\textbf{82}\pm3$  \\
Dense   &                                                                    & $10\pm0$ & $14\pm2$ & $24\pm8$  & $\textbf{43}\pm5$  \\ \bottomrule
\end{tabular}
\end{table}

It can be seen from the Table \ref{table1} that our proposed model has significantly improved the performance and robustness of the end-to-end driving network, achieving a 100\% success rate in the static navigation tasks and a better success rate than the prior models in the dynamic navigation tasks. The better generalization capability of our model can be manifested in the testing scenario where both the road layout and weather conditions are different from that of the training conditions. As listed in the Table \ref{table1}, in the more challenging testing scenario, there is no decline in the success rate in the static navigation tasks, and only a slight drop is observed in success rate when navigating in dynamic traffic. The proposed method, as seen in Table \ref{table2}, still outperforms the prior models in the more challenging NoCrash benchmark, and no reduction in success rate is found when the end-to-end agent fulfills the navigation tasks in empty traffic. It is expected that the agent performs poorly in dense traffic because we did not train the model to obey the traffic rules (e.g. stop for red light), whereas our model still demonstrates better performance over others.

\begin{figure}[htp]
    \centering
    
    \subfloat[]{\includegraphics[width=0.47\linewidth]{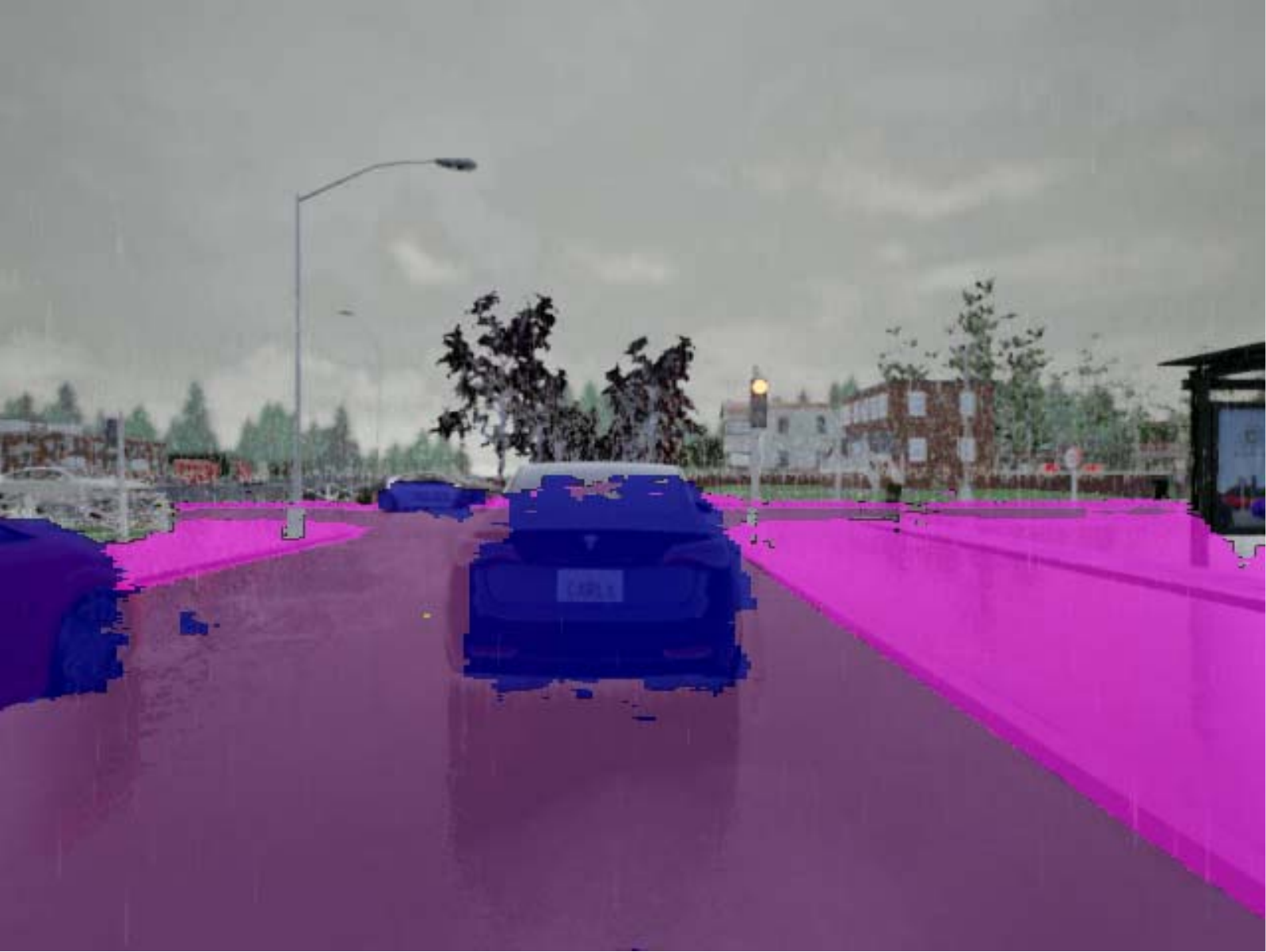}%
    \label{fig4a}}
    \hfill
    \subfloat[]{\includegraphics[width=0.47\linewidth]{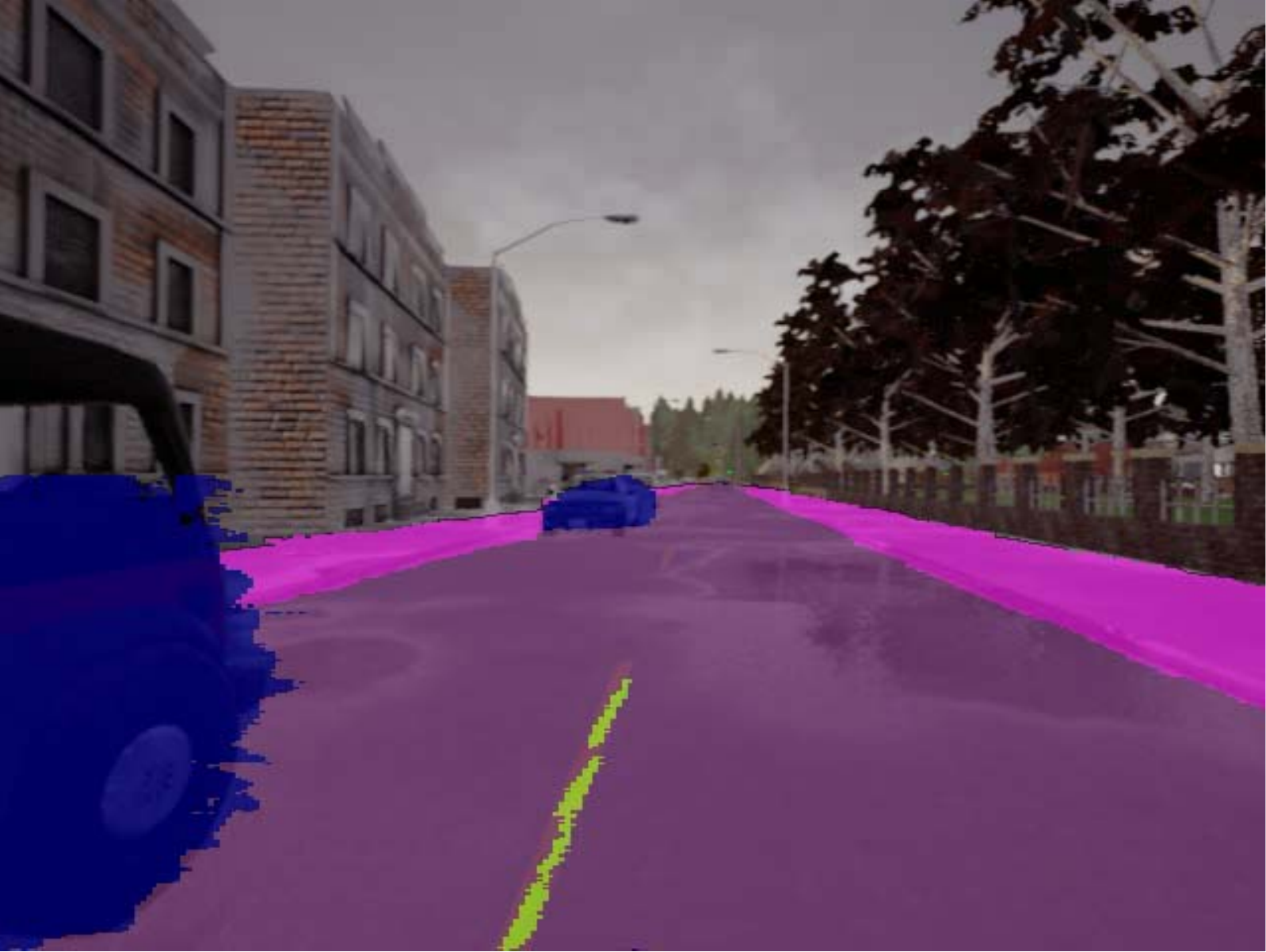}%
    \label{fig4b}}
    
    \subfloat[]{\includegraphics[width=0.47\linewidth]{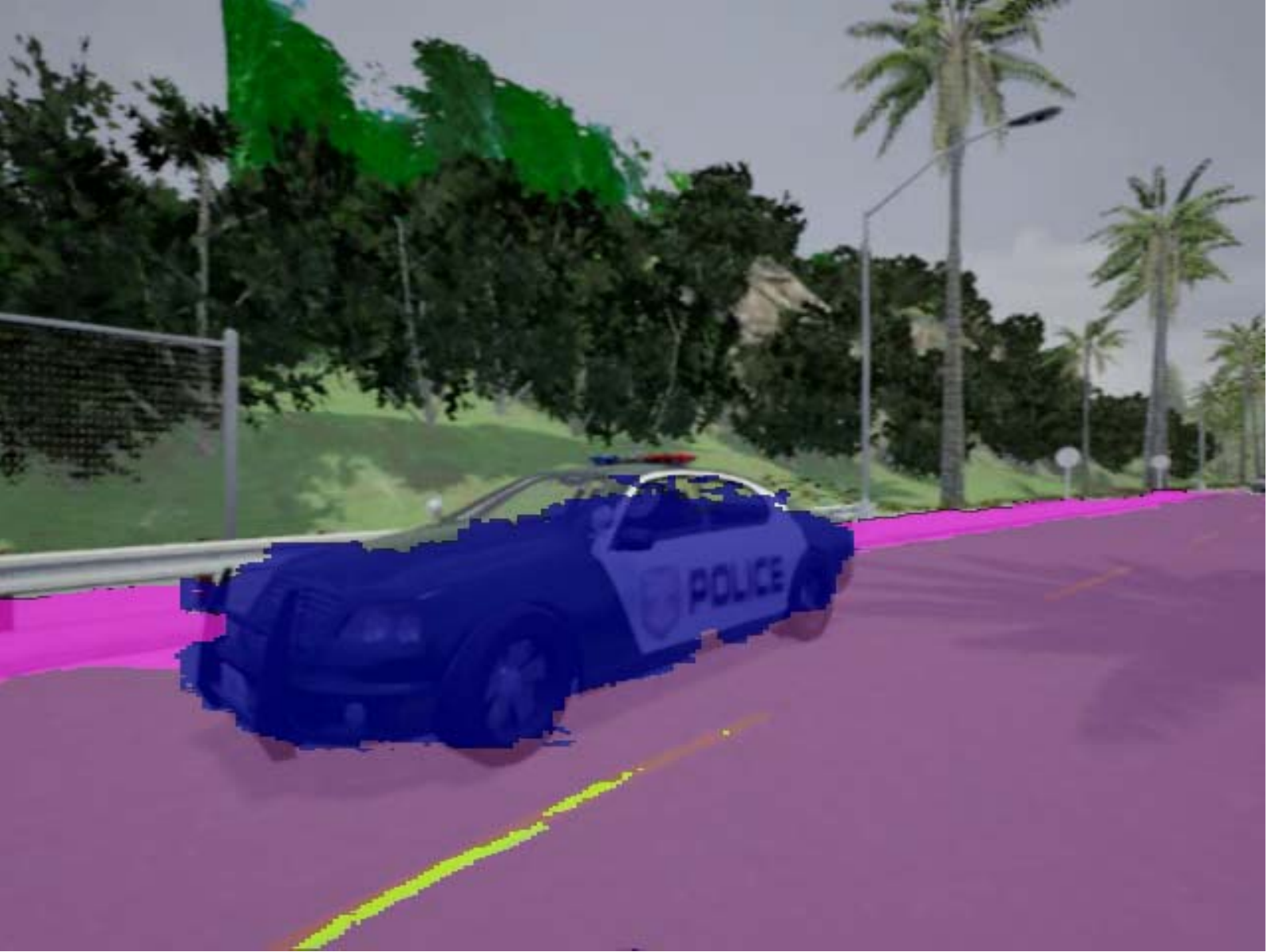}%
    \label{fig4c}}
    \hfill
    \subfloat[]{\includegraphics[width=0.47\linewidth]{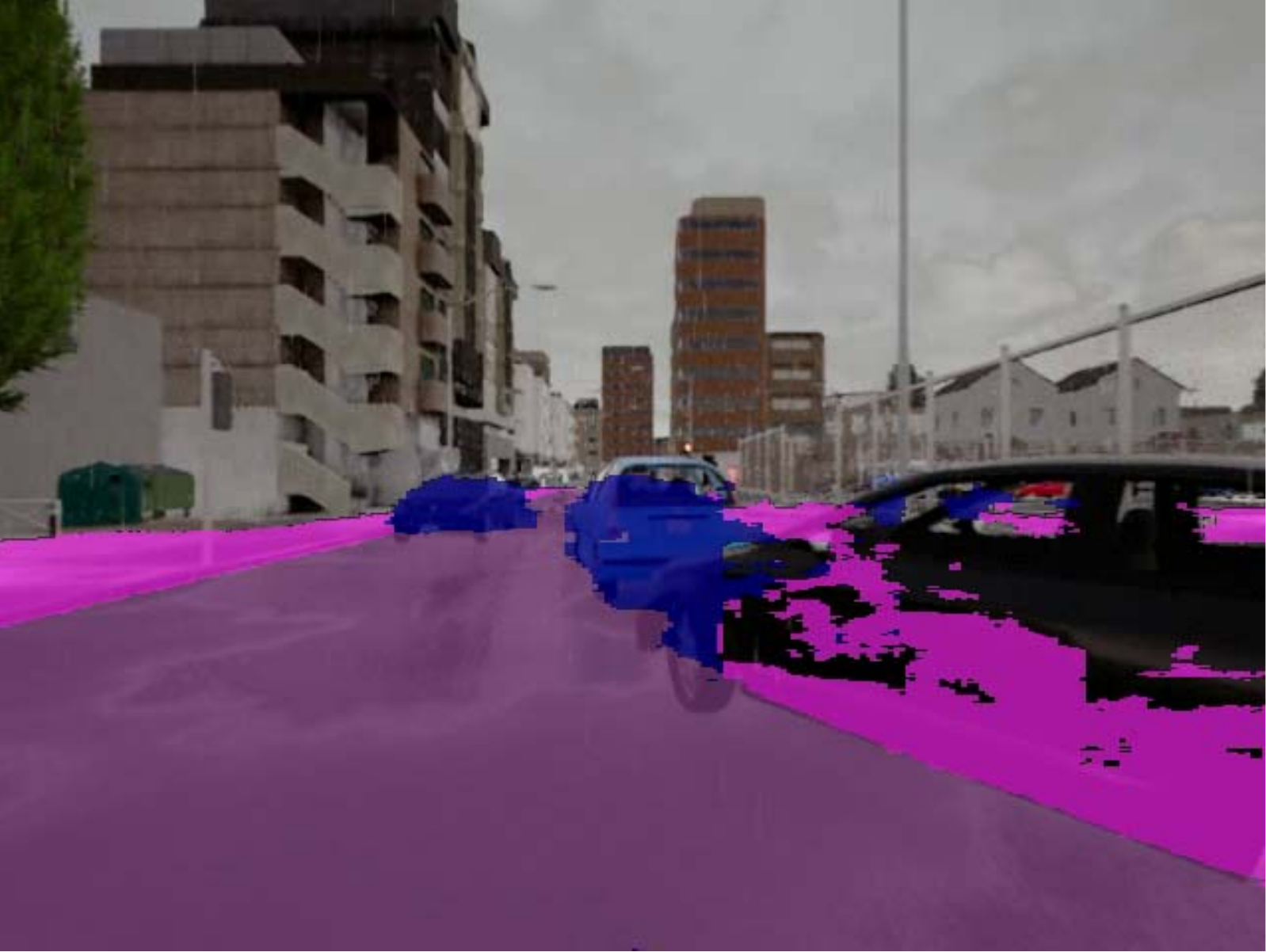}%
    \label{fig4d}}
    
    \caption{Demonstrations of model performance: (a) collision avoidance; (b) error correction; (c) failure case with correct scene understanding; (d) failure case with false scene understanding.}
    \label{fig4}
\end{figure}

Here, some typical cases of the scene understanding capability of our model and some failure cases are presented in Fig. \ref{fig4}. Fig. \ref{fig4a} shows a scenario of collision avoidance, in which our model is able to recognize the feasible driving area, sidewalk, and other vehicles and output an appropriate braking command to stop the vehicle behind the obstacle. Fig. \ref{fig4b} shows that the end-to-end agent is attempting to correct its deviation and get back to the lane center. These capabilities keep the agent staying on the road and avoiding collisions with other obstacles, achieving a 100\% success rate in the static navigation tasks and a higher success rate in the dynamic navigation tasks, compared to other ones. The failures are mostly due to collisions with surrounding vehicles or pedestrians. The main reason for these failure cases is because that the encountered scenes are out of the training data distribution. Fig. \ref{fig4c} and Fig. \ref{fig4d} show two failure cases. Fig. \ref{fig4c} shows a scenario of a crash in a sharp turn. Although it correctly recognizes the vehicle ahead, the model fails to generate an appropriate steering control. This failure can be attributed to that the driving policy has never been trained in this scene. In Fig \ref{fig4d}, the model fails to identify the vehicle on the right side and thus causes a crash. In general, apart from some out-of-distribution cases, our model has shown good generalization capability, resulting in a higher success rate in unobserved urban situations.

\subsection{Ablation study}
\begin{figure}[htp]
    \centering
    
    \subfloat[]{\includegraphics[width=\linewidth]{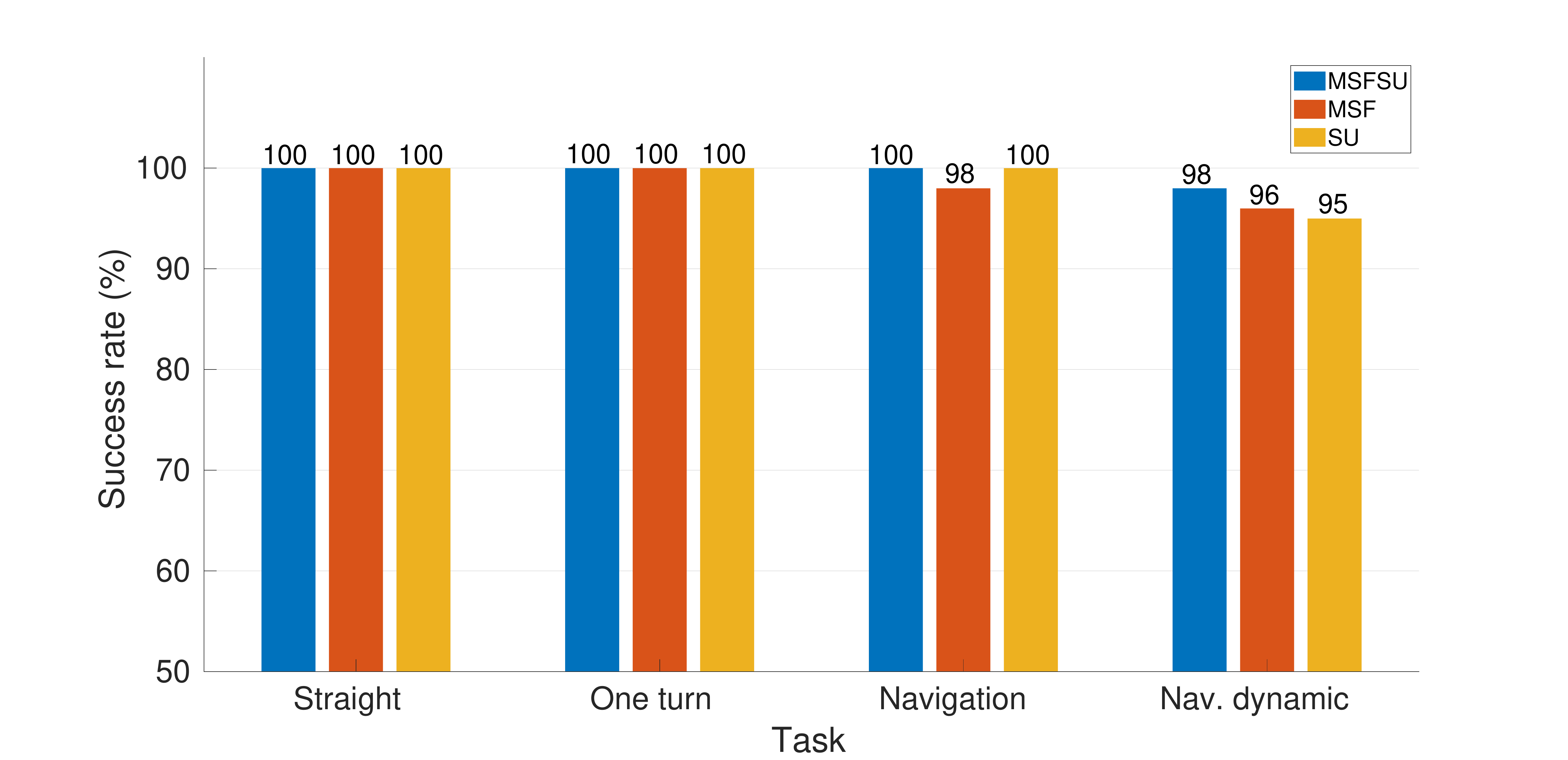}}%
    \label{fig5a}
    \hfill
    \subfloat[]{\includegraphics[width=\linewidth]{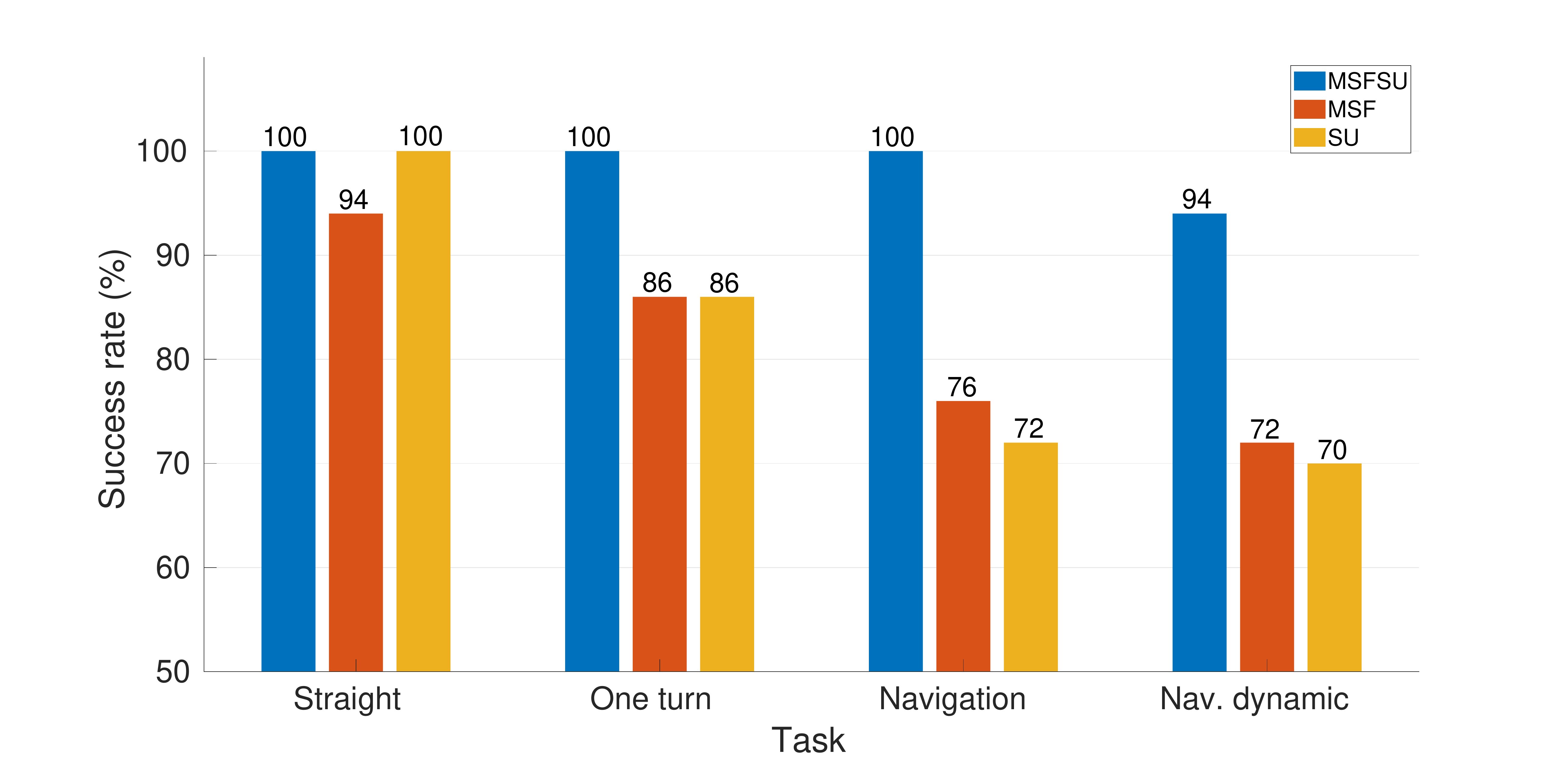}}%
    \label{fig5b}
    
    \caption{Results of success rate on the CoRL2017 benchmark: (a) Training scenario; (b) Testing Scenario.}
    \label{fig5}
\end{figure}

The performance improvement of the proposed end-to-end driving approach may be contributed by several factors, e.g. multimodal sensor fusion, scene understanding, and/or their combined effect. Therefore, this ablation test is designed to investigate the impacts of multimodal sensor fusion and scene understanding on the performance of the developed model. First, we remove the depth information from the network inputs with maintaining the scene understanding decoder to perform semantic segmentation on the input image. Here we denote it as scene understanding (SU) model. The system is similar to the multi-task learning model \cite{li2018rethinking} with the removal of the depth estimation. Next, we take the scene understanding decoder away from the network and the associated semantic segmentation task, and thereby the system is similar to the multimodal early fusion model \cite{xiao2019multimodal}. We denote this model as the multimodal sensor fusion (MSF) model. We evaluate their performance regarding the success rate on the CoRL2017 benchmark, and the results are illustrated in Fig. \ref{fig5}.

\begin{figure}[htp]
    \centering
    
    \subfloat[]{\includegraphics[width=0.47\linewidth]{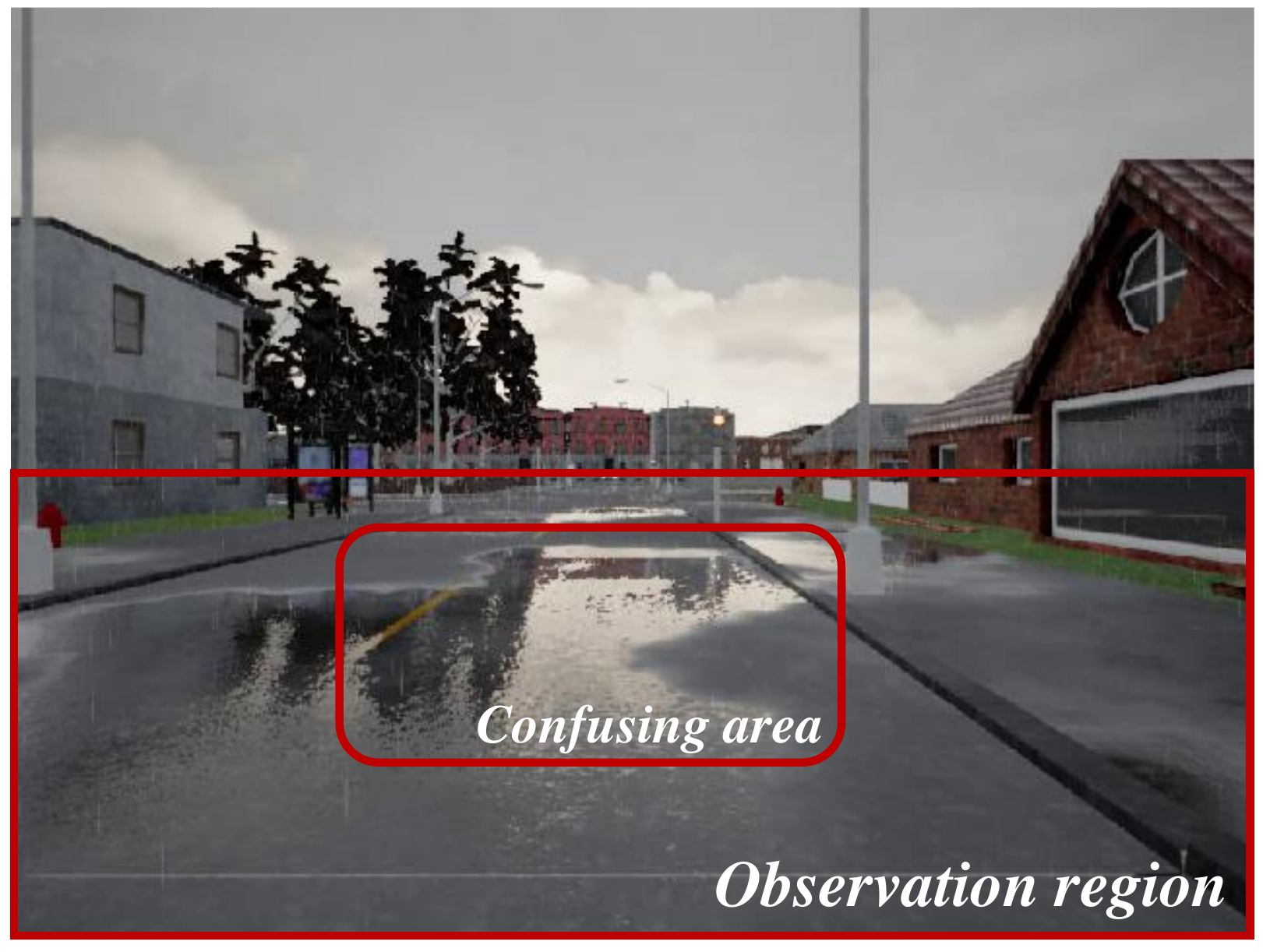}%
    \label{fig6a}}
    \hfil
    \subfloat[]{\includegraphics[width=0.47\linewidth]{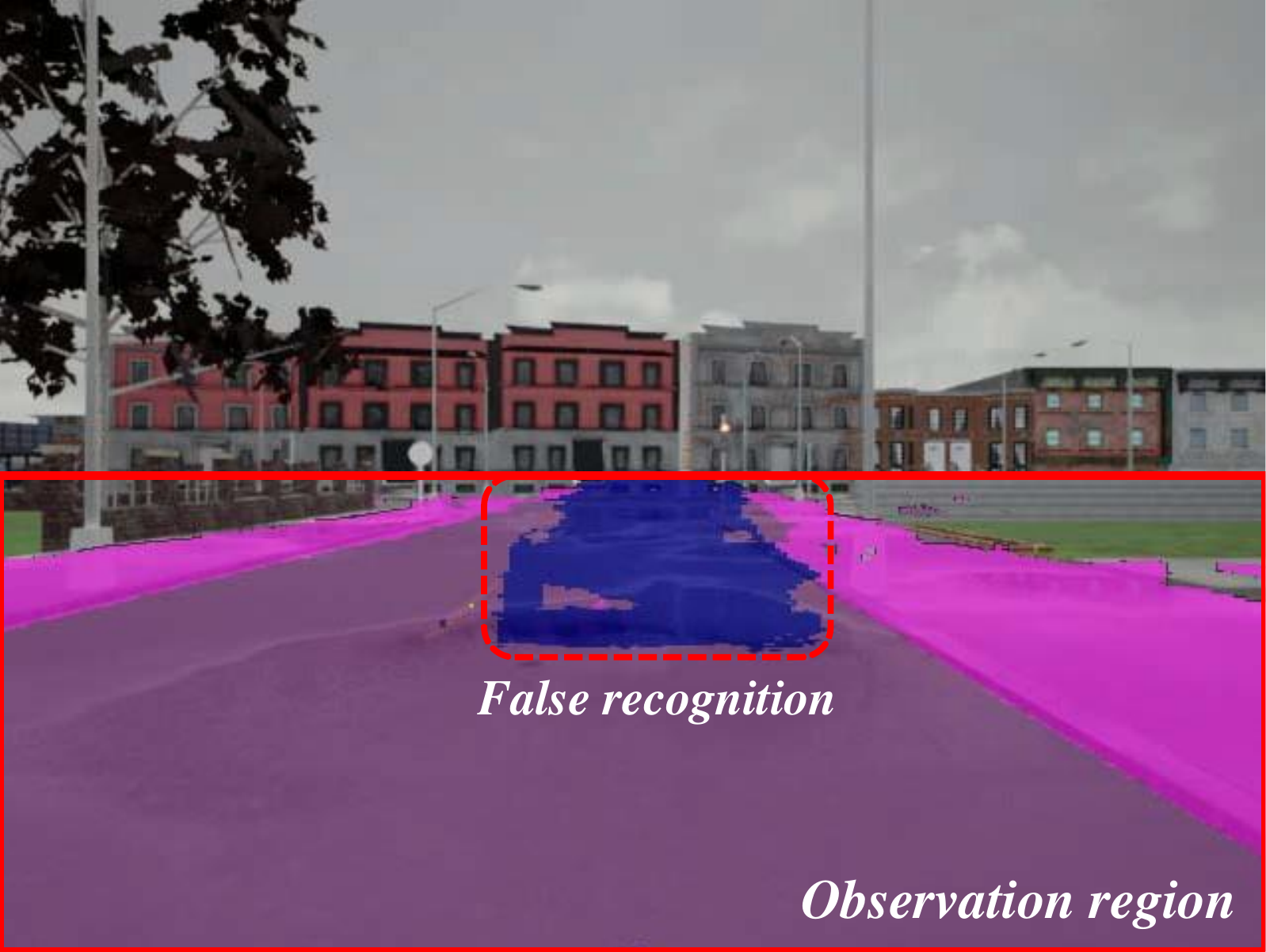}%
    \label{fig6b}}

    \caption{Failure cases of unexpected stops from different models: (a) the multimodal sensor fusion model; (b) the scene understanding model.}
    \label{fig6}
\end{figure}

The results in Fig. \ref{fig5} indicate that both multimodal sensor fusion and scene understanding play essential roles in the developed end-to-end autonomous driving model. The models without multimodal sensor fusion or scene understanding can perform approximately the same as the original one in the training scenario. However, their generalization performance is significantly impaired, consequently resulting in a substantial decrease in success rate in navigation tasks under unobserved situations. In the static navigation tasks under the testing scenario, most of the failure cases in the MSF model and SU model are caused by unexpected stops in unobserved weather settings. As shown in Fig. \ref{fig6a}, the MSF model outputs a false brake command to stop the vehicle in front of a puddle. A possible reason is that the standing water on the road may confuse the network to regard it as an obstacle ahead. Although we have added depth information, the MSF model still makes a wrong judgment. This phenomenon happens in Fig. \ref{fig6b}, where the SU model makes a false recognition that there is a vehicle in the front. Nevertheless, the proposed model combining multimodal sensor fusion and scene understanding can avoid such a mistake caused by false perception. This finding suggests that the better performance of our model is contributed by the combined effect of multimodal sensor fusion and scene understanding. On one hand, incorporating scene understanding capability could help the network find more relevant and general features of the driving scene. One the other hand, adding depth information could further enhance the scene understanding ability.

\begin{figure*}[htp]
    \centering
    
    \subfloat[]{\includegraphics[width=0.48\linewidth]{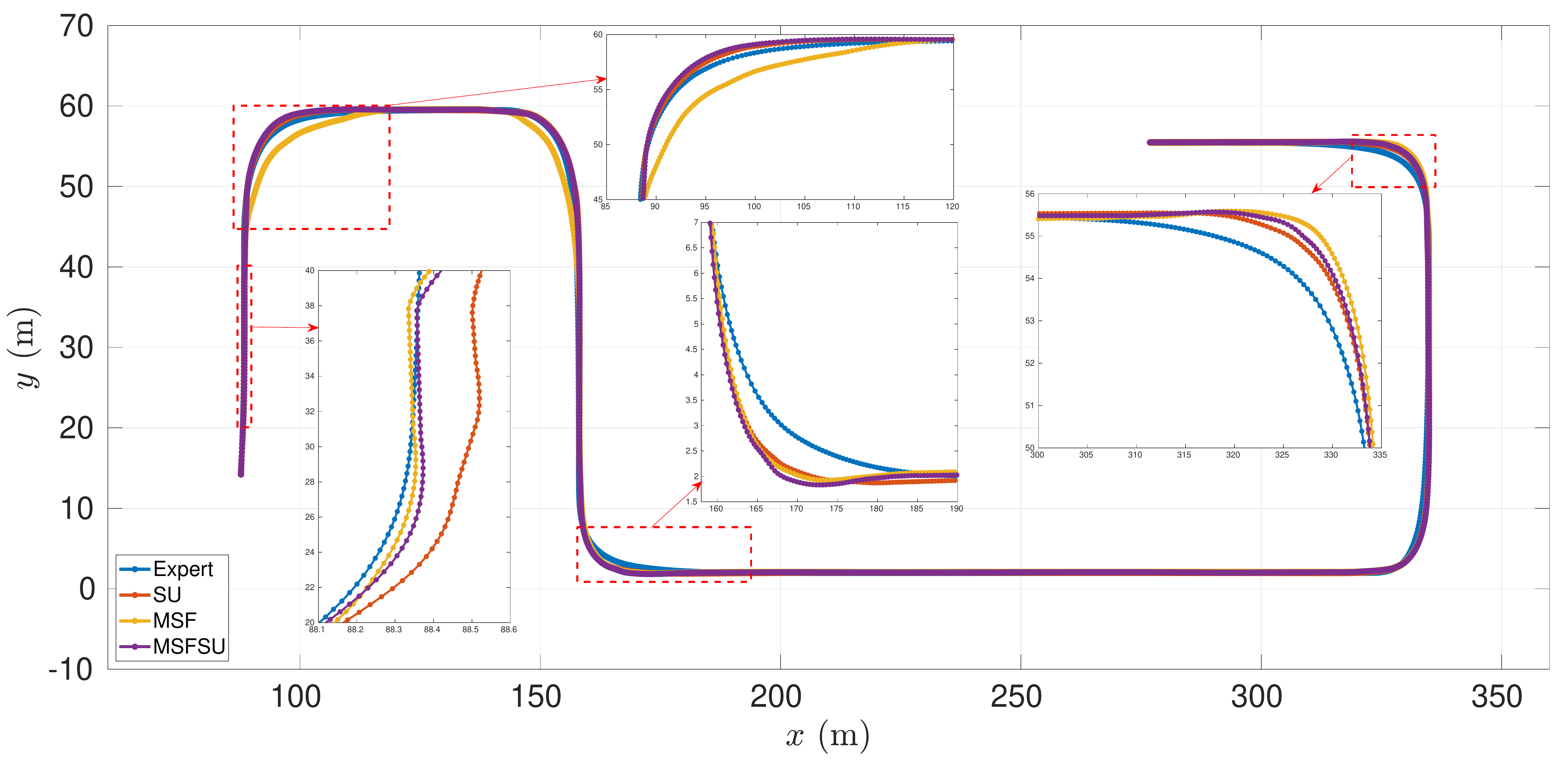}%
    \label{fig7a}}
    \hfil
    \subfloat[]{\includegraphics[width=0.48\linewidth]{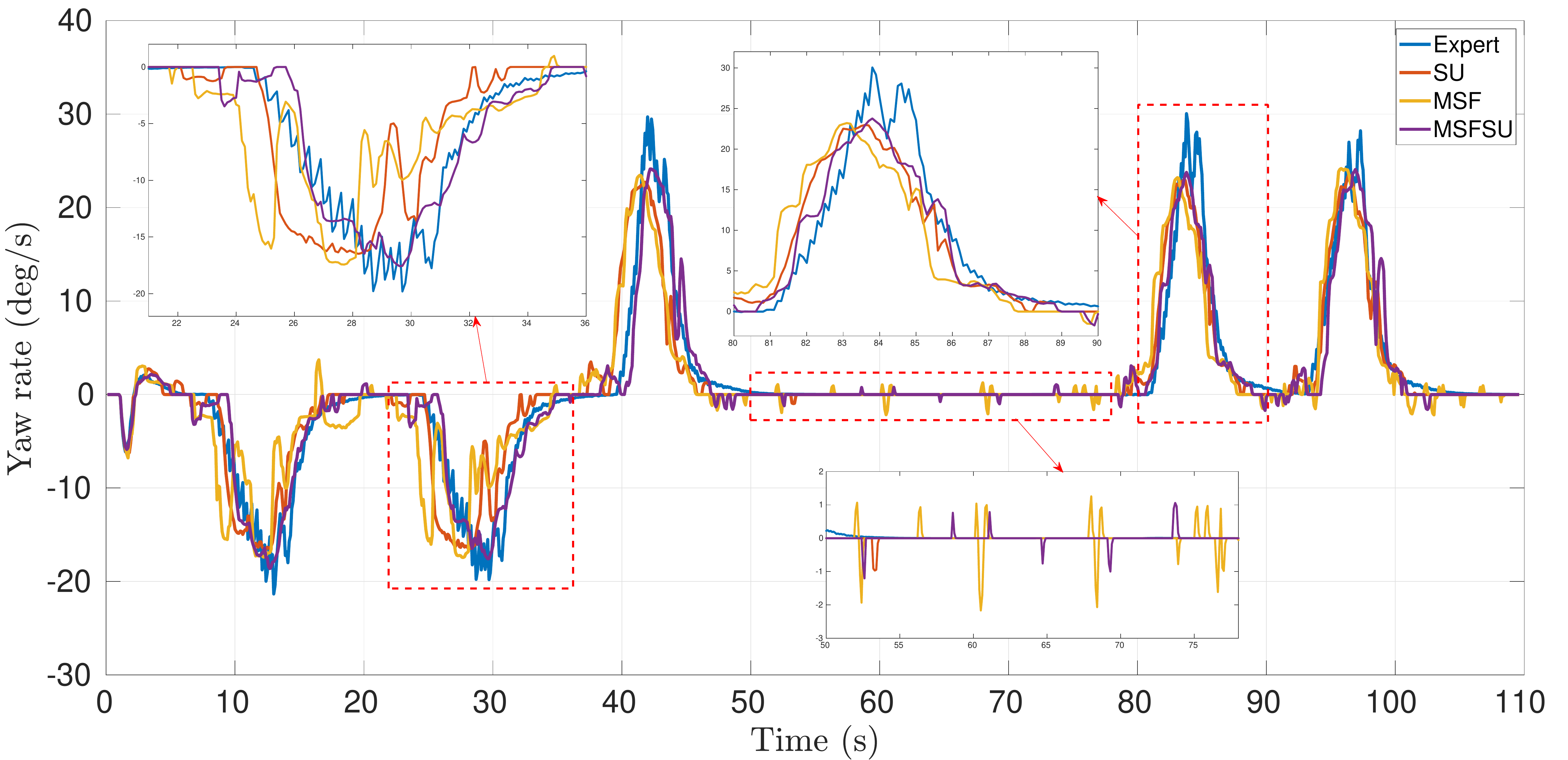}%
    \label{fig7b}}

    \subfloat[]{\includegraphics[width=0.48\linewidth]{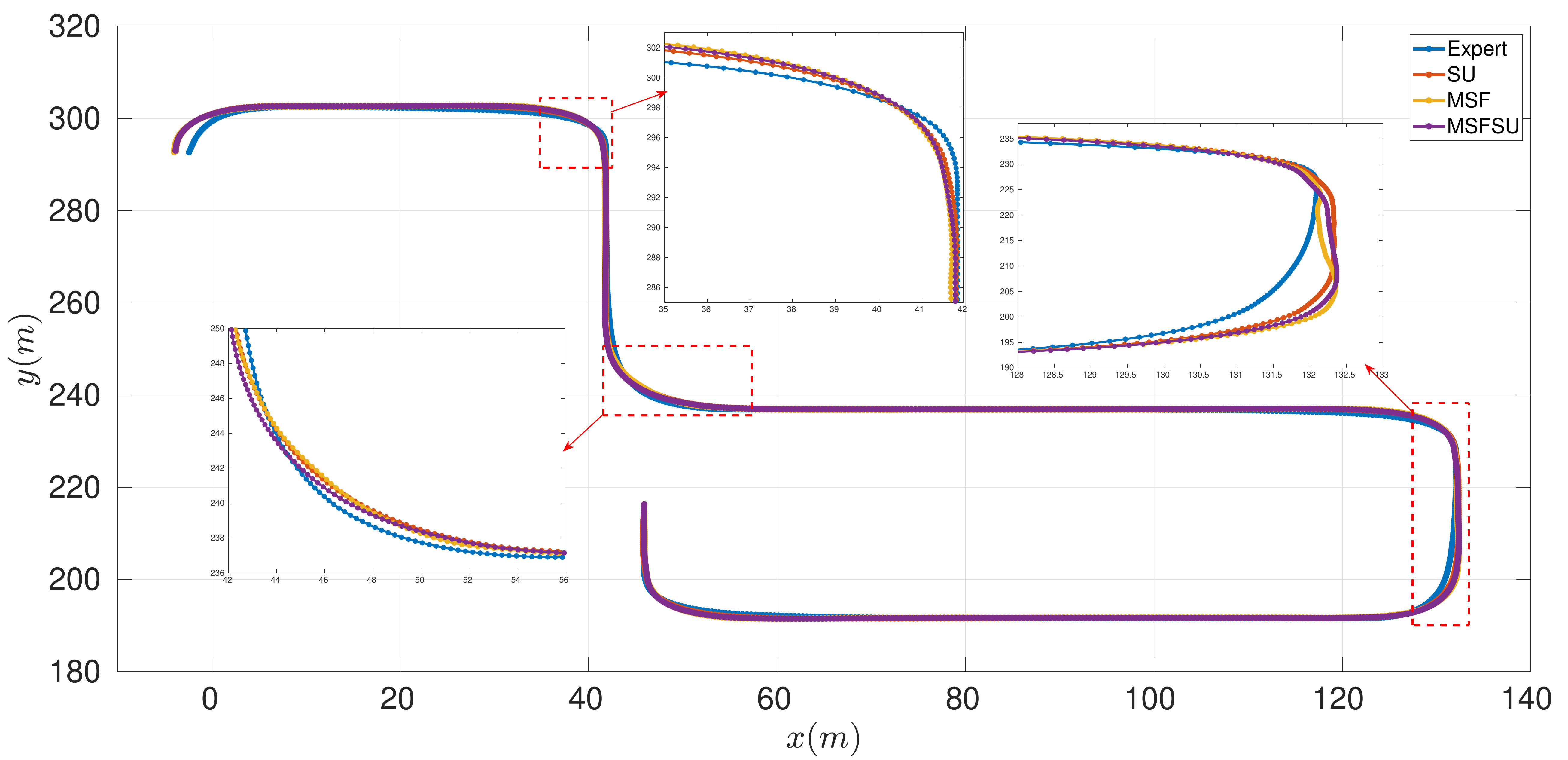}%
    \label{fig7c}}
    \hfil
    \subfloat[]{\includegraphics[width=0.48\linewidth]{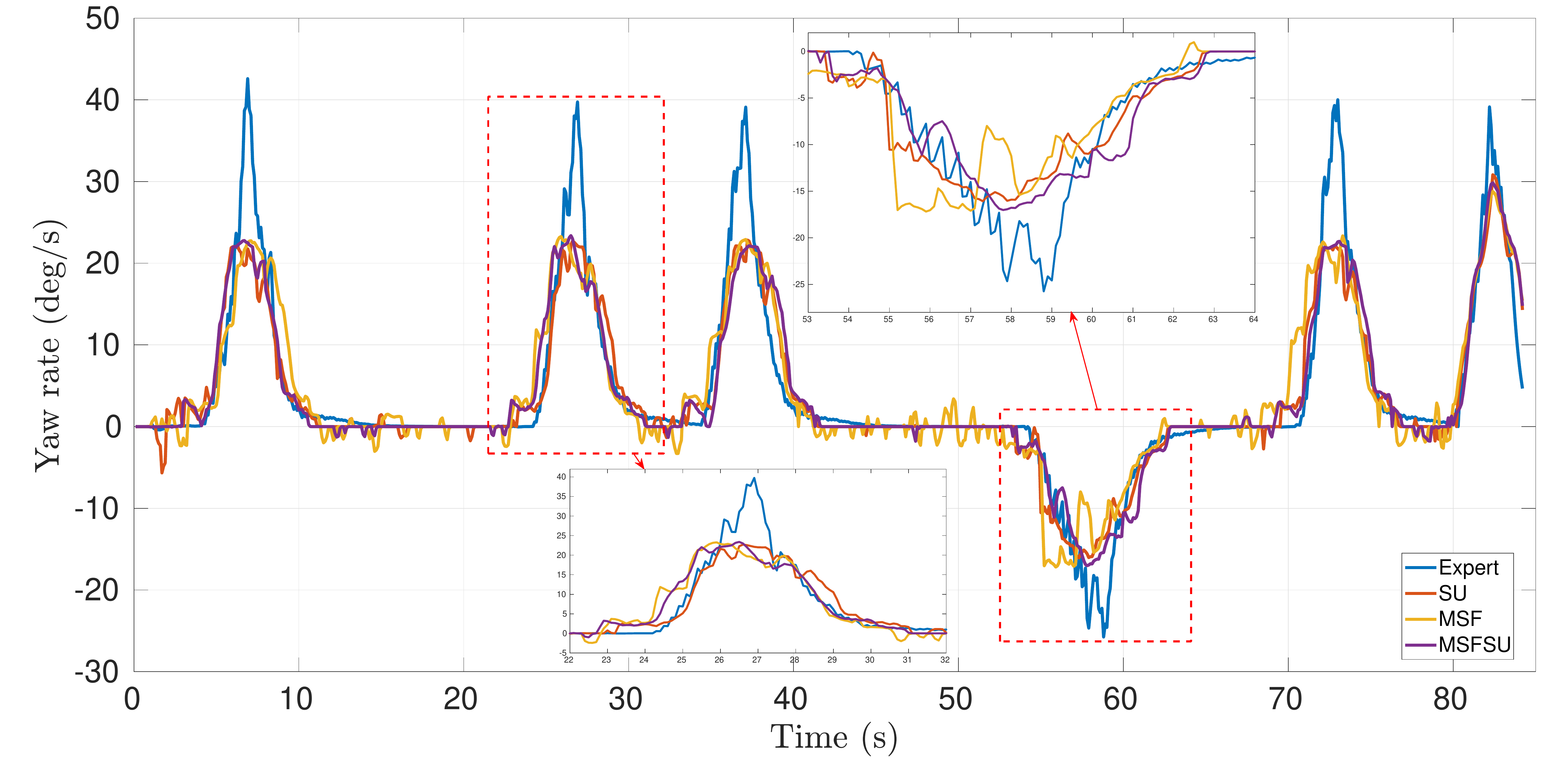}%
    \label{fig7d}}
    
    \caption{Examples of trajectories and yaw rates of different agents: (a) the trajectory of an episode in the training scenario; (b) the yaw rate of an episode in the training scenario; (c) the trajectory of an episode in the testing scenario; (d) the yaw rate of an episode in the testing scenario.}
    \label{fig7}
\end{figure*}

Then, we compare the three models in terms of their closeness to the expert demonstrations. The evaluation metric is defined as the root mean square error (RMSE) in meters between the trajectory of the end-to-end agent and that of the expert over a specific task: $RMSE = \frac{1}{{{N_e}}}\sum\limits_{i = 1}^{{N_e}} {\left( {\sqrt {\frac{1}{T}\sum\limits_{t = 1}^T {\left( {{{\left( {{x_{it}} - x_{it}^E} \right)}^2} + {{\left( {{y_{it}} - y_{it}^E} \right)}^2}} \right)} } } \right)}$, where $N_e$ is the number of successful episodes in a task and $T$ is the number of timesteps to finish an episode; ${x_{it}^E}$ and ${y_{it}^E}$ are on the trajectory of the demonstrations, while ${{x_{it}}}$ and ${{y_{it}}}$ are on the the trajectory of the end-to-end agent. The expert demonstrations are collected by using the built-in autopilot function in the CARLA simulator. We evaluate these models on the static navigation tasks in the CoRL2017 benchmark, and the results are listed in Table \ref{table3}.

\begin{table}[htp]
\centering
\caption{Comparison of RMSE on the static navigation tasks in the CoRL2017 benchmark}
\label{table3}
\begin{tabular}{@{}cccccc@{}}
\toprule
Task       & Scenario                                                          & MSF     & SU              & MSFSU           \\ \midrule
Straight   & \multirow{3}{*}{\begin{tabular}[c]{@{}c@{}}Training \\scenario\end{tabular}} & 0.1288  & 0.1155          & \textbf{0.1131} \\
One turn   &                                                                   & 0.2809  & \textbf{0.2504} & 0.2572          \\
Navigation &                                                                   & 0.2242  & 0.1971          & \textbf{0.1970} \\ \midrule
Straight   & \multirow{3}{*}{\begin{tabular}[c]{@{}c@{}}Testing\\ scenario\end{tabular}}  & 0.1313  & 0.1217          & \textbf{0.1144} \\
One turn   &                                                                   & 0.3603 & \textbf{0.2921} & 0.3167           \\
Navigation &                                                                   & 0.2965 & \textbf{0.2871} & 0.2897           \\ \bottomrule
\end{tabular}
\end{table}

The results in Table \ref{table3} indicate that the end-to-end driving models with scene understanding could get closer to the expert demonstrations, and the proposed MSFSU model behaves the best in the lane-keeping task. In contrast, the model with only multimodal sensor fusion has a more significant deviation from expert demonstrations, especially in the turning task. We display two examples in Fig. \ref{fig7}, showing the trajectories of different agents with their corresponding yaw rates of an episode in the training scenario and testing scenario, respectively.

We notice from Fig. \ref{fig7} that in general, the trajectories of all end-to-end agents are consistent with that of the expert demonstration. However, although there exist some deviations during turning, the end-to-end agents are able to get back onto the correct track with a lower value of peak yaw rate. The end-to-end driving model without scene understanding has more significant deviations than others, as shown in Fig. \ref{fig7a}. In terms of the yaw rate, the proposed MSFSU model possesses a smoother yaw rate curve, as reflected in Fig. \ref{fig7b}. While the MSF model needs to correct its direction continually, causing more oscillations in the yaw rate. The above results validate the reasonable control quality of the proposed MSFSU model.

\subsection{Discussions}

Some recent works \cite{chen2019learning, zhao2019lates} have achieved similar results to our method on the CoRL2017 benchmark. However, they utilize more privileged information and sophisticated training methods. \cite{chen2019learning} utilizes maps and ground-truth information of all traffic participants to train a privileged agent, which is later used for online training of a vision-based agent. \cite{zhao2019lates} first trains an expert agent that leverages side information on semantics and stop intentions, and then a student agent that tries to distill the latent feature of the expert agent but only receives images as input. On the contrary, only the semantic segmentation image is used as privileged information in our model, and the depth information can be obtained through perception sensors. Moreover, our model is trained purely offline with a simpler training procedure but still reaches a comparative performance to these sophisticated models.

However, there are still some issues and limitations that need to be further improved in the future. The most significant one is the simulation-to-real transfer. The difficulty of the transfer is primarily caused by three factors. First, the texture, illumination, and details of the camera images from the simulator are quite different from that in the real world, and the depth-sensing cannot be exactly precise in the real world. Therefore, the end-to-end driving network may not function well under the domain shift from simulation to the real world. Second, labelling the visual images with semantics for network training in the real world is laborious and time consuming. Last but not least, the computational burden becomes a major concern for onboard application in a real vehicle.

Many other works have added ego-speed as an input modality to the end-to-end network, however, this does not work in our model. Adding feedback speed through a fully connected network, or adding the feedback speed sequence through a long-short term memory network directly to the driving policy, has caused the inertia problem, i.e. the agent cannot restart after it stops for obstacle avoidance. It is because the network fully relies on the feedback speed information to reduce the training loss in the speed prediction, but the system can barely work in the online deployment. This further verifies our point of view that multimodal information itself does not guarantee a better result. Therefore, we should investigate the mechanisms of incorporating the speed information in a scene understanding task, and this is one direction of our future work. Furthermore, future research should concentrate on other scene understanding tasks such as restoring the images of the driving scene from the latent representation using an autoencoder, or predicting the short-term change of the image based on motion and temporal information. Other more advanced and abstract scene understanding representations, such as constructing bird-view maps from on-board sensors, detecting objects and lanes, or even learning traffic rules, are also worthwhile exploring. 

\section{Conclusions}
\label{sec5}
In this study, we propose a novel structure of deep neural networks for end-to-end autonomous driving. The developed network consists of multimodal sensor fusion, scene understanding, and conditional driving policy modules. The multimodal sensor fusion encoder fuses the visual image and depth information into a low-dimension latent representation, followed by the scene understanding decoder to perform pixel-wise semantic segmentation. The conditional driving policy outputs vehicle control commands simultaneously. We test and evaluate our model in the CARLA simulator with the CoRL2017 and NoCrash benchmarks. Testing results show that the proposed model is advantageous over other existing ones in terms of task success rate and generalization ability, achieving a 100\% success rate in static navigation tasks in both training and unobserved circumstances. The further ablation study finds out that the performance improvement of our model is contributed by the combined effects of multimodal sensor fusion and scene understanding. The models without multimodal sensor fusion or scene understanding suffer from the drop in success rate by 28\% and 22\%, respectively, on the navigation task in the unobserved environment. Furthermore, our end-to-end driving model behaves more closely to the expert demonstrations with the RMSE of 0.1970 meters compared to the multimodal sensor fusion model with the RMSE of 0.2242 meters on the navigation task in the training scenario, presenting a better control quality. The testing results and performance comparison demonstrate the feasibility and effectiveness of the developed deep neural network for end-to-end autonomous driving.

\bibliographystyle{IEEEtran}
\bibliography{ref.bib}

\begin{thebibliography}{10}
\providecommand{\url}[1]{#1}
\csname url@samestyle\endcsname
\providecommand{\newblock}{\relax}
\providecommand{\bibinfo}[2]{#2}
\providecommand{\BIBentrySTDinterwordspacing}{\spaceskip=0pt\relax}
\providecommand{\BIBentryALTinterwordstretchfactor}{4}
\providecommand{\BIBentryALTinterwordspacing}{\spaceskip=\fontdimen2\font plus
\BIBentryALTinterwordstretchfactor\fontdimen3\font minus
  \fontdimen4\font\relax}
\providecommand{\BIBforeignlanguage}[2]{{%
\expandafter\ifx\csname l@#1\endcsname\relax
\typeout{** WARNING: IEEEtran.bst: No hyphenation pattern has been}%
\typeout{** loaded for the language `#1'. Using the pattern for}%
\typeout{** the default language instead.}%
\else
\language=\csname l@#1\endcsname
\fi
#2}}
\providecommand{\BIBdecl}{\relax}
\BIBdecl

\bibitem{schwarting2018planning}
W.~Schwarting, J.~Alonso-Mora, and D.~Rus, ``Planning and decision-making for
  autonomous vehicles,'' \emph{Annual Review of Control, Robotics, and
  Autonomous Systems}, 2018.

\bibitem{bojarski2016end}
M.~Bojarski, D.~Del~Testa, D.~Dworakowski, B.~Firner, B.~Flepp, P.~Goyal, L.~D.
  Jackel, M.~Monfort, U.~Muller, J.~Zhang \emph{et~al.}, ``End to end learning
  for self-driving cars,'' \emph{arXiv preprint arXiv:1604.07316}, 2016.

\bibitem{chen2017end}
Z.~Chen and X.~Huang, ``End-to-end learning for lane keeping of self-driving
  cars,'' in \emph{2017 IEEE Intelligent Vehicles Symposium (IV)}.\hskip 1em
  plus 0.5em minus 0.4em\relax IEEE, 2017, pp. 1856--1860.

\bibitem{kendall2019learning}
A.~Kendall, J.~Hawke, D.~Janz, P.~Mazur, D.~Reda, J.-M. Allen, V.-D. Lam,
  A.~Bewley, and A.~Shah, ``Learning to drive in a day,'' in \emph{2019
  International Conference on Robotics and Automation (ICRA)}.\hskip 1em plus
  0.5em minus 0.4em\relax IEEE, 2019, pp. 8248--8254.

\bibitem{hawke2019urban}
J.~Hawke, R.~Shen, C.~Gurau, S.~Sharma, D.~Reda, N.~Nikolov, P.~Mazur,
  S.~Micklethwaite, N.~Griffiths, A.~Shah \emph{et~al.}, ``Urban driving with
  conditional imitation learning,'' \emph{arXiv preprint arXiv:1912.00177},
  2019.

\bibitem{yang2017feature}
S.~Yang, W.~Wang, C.~Liu, W.~Deng, and J.~K. Hedrick, ``Feature analysis and
  selection for training an end-to-end autonomous vehicle controller using deep
  learning approach,'' in \emph{2017 IEEE Intelligent Vehicles Symposium
  (IV)}.\hskip 1em plus 0.5em minus 0.4em\relax IEEE, 2017, pp. 1033--1038.

\bibitem{bojarski2017explaining}
M.~Bojarski, P.~Yeres, A.~Choromanska, K.~Choromanski, B.~Firner, L.~Jackel,
  and U.~Muller, ``Explaining how a deep neural network trained with end-to-end
  learning steers a car,'' \emph{arXiv preprint arXiv:1704.07911}, 2017.

\bibitem{codevilla2019exploring}
F.~Codevilla, E.~Santana, A.~M. L{\'o}pez, and A.~Gaidon, ``Exploring the
  limitations of behavior cloning for autonomous driving,'' in
  \emph{Proceedings of the IEEE International Conference on Computer Vision},
  2019, pp. 9329--9338.

\bibitem{xiao2019multimodal}
Y.~Xiao, F.~Codevilla, A.~Gurram, O.~Urfalioglu, and A.~M. L{\'o}pez,
  ``Multimodal end-to-end autonomous driving,'' \emph{arXiv preprint
  arXiv:1906.03199}, 2019.

\bibitem{amado2019end}
J.~A.~D. Amado, I.~P. Gomes, J.~Amaro, D.~F. Wolf, and F.~S. Os{\'o}rio,
  ``End-to-end deep learning applied in autonomous navigation using
  multi-cameras system with rgb and depth images,'' in \emph{2019 IEEE
  Intelligent Vehicles Symposium (IV)}.\hskip 1em plus 0.5em minus 0.4em\relax
  IEEE, 2019, pp. 1626--1631.

\bibitem{sobh2018end}
I.~Sobh, L.~Amin, S.~Abdelkarim, K.~Elmadawy, M.~Saeed, O.~Abdeltawab,
  M.~Gamal, and A.~El~Sallab, ``End-to-end multi-modal sensors fusion system
  for urban automated driving,'' \emph{NIPS 2018 Workshop MLITS}, 2018.

\bibitem{chen2018lidar}
Y.~Chen, J.~Wang, J.~Li, C.~Lu, Z.~Luo, H.~Xue, and C.~Wang, ``Lidar-video
  driving dataset: Learning driving policies effectively,'' in
  \emph{Proceedings of the IEEE Conference on Computer Vision and Pattern
  Recognition}, 2018, pp. 5870--5878.

\bibitem{xu2017end}
H.~Xu, Y.~Gao, F.~Yu, and T.~Darrell, ``End-to-end learning of driving models
  from large-scale video datasets,'' in \emph{Proceedings of the IEEE
  conference on computer vision and pattern recognition}, 2017, pp. 2174--2182.

\bibitem{li2018rethinking}
Z.~Li, T.~Motoyoshi, K.~Sasaki, T.~Ogata, and S.~Sugano, ``Rethinking
  self-driving: Multi-task knowledge for better generalization and accident
  explanation ability,'' \emph{arXiv preprint arXiv:1809.11100}, 2018.

\bibitem{wang2019deep}
D.~Wang, C.~Devin, Q.-Z. Cai, F.~Yu, and T.~Darrell, ``Deep object-centric
  policies for autonomous driving,'' in \emph{2019 International Conference on
  Robotics and Automation (ICRA)}.\hskip 1em plus 0.5em minus 0.4em\relax IEEE,
  2019, pp. 8853--8859.

\bibitem{badrinarayanan2017segnet}
V.~Badrinarayanan, A.~Kendall, and R.~Cipolla, ``Segnet: A deep convolutional
  encoder-decoder architecture for image segmentation,'' \emph{IEEE
  transactions on pattern analysis and machine intelligence}, vol.~39, no.~12,
  pp. 2481--2495, 2017.

\bibitem{codevilla2018end}
F.~Codevilla, M.~Miiller, A.~L{\'o}pez, V.~Koltun, and A.~Dosovitskiy,
  ``End-to-end driving via conditional imitation learning,'' in \emph{2018 IEEE
  International Conference on Robotics and Automation (ICRA)}.\hskip 1em plus
  0.5em minus 0.4em\relax IEEE, 2018, pp. 1--9.

\bibitem{yuan2019steeringloss}
W.~Yuan, M.~Yang, C.~Wang, and B.~Wang, ``Steeringloss: Theory and application
  for steering prediction,'' in \emph{2019 IEEE Intelligent Vehicles Symposium
  (IV)}.\hskip 1em plus 0.5em minus 0.4em\relax IEEE, 2019, pp. 1420--1425.

\bibitem{he2016identity}
K.~He, X.~Zhang, S.~Ren, and J.~Sun, ``Identity mappings in deep residual
  networks,'' in \emph{European conference on computer vision}.\hskip 1em plus
  0.5em minus 0.4em\relax Springer, 2016, pp. 630--645.

\bibitem{dosovitskiy2017carla}
A.~Dosovitskiy, G.~Ros, F.~Codevilla, A.~Lopez, and V.~Koltun, ``Carla: An open
  urban driving simulator,'' \emph{arXiv preprint arXiv:1711.03938}, 2017.

\bibitem{liang2018cirl}
X.~Liang, T.~Wang, L.~Yang, and E.~Xing, ``Cirl: Controllable imitative
  reinforcement learning for vision-based self-driving,'' in \emph{Proceedings
  of the European Conference on Computer Vision (ECCV)}, 2018, pp. 584--599.

\bibitem{sauer2018conditional}
A.~Sauer, N.~Savinov, and A.~Geiger, ``Conditional affordance learning for
  driving in urban environments,'' \emph{arXiv preprint arXiv:1806.06498},
  2018.

\bibitem{chen2019learning}
D.~Chen, B.~Zhou, V.~Koltun, and P.~Kr{\"a}henb{\"u}hl, ``Learning by
  cheating,'' \emph{arXiv preprint arXiv:1912.12294}, 2019.

\bibitem{zhao2019lates}
A.~Zhao, T.~He, Y.~Liang, H.~Huang, G.~V.~d. Broeck, and S.~Soatto, ``Lates:
  Latent space distillation for teacher-student driving policy learning,''
  \emph{arXiv preprint arXiv:1912.02973}, 2019.

\end{thebibliography}

\begin{IEEEbiography}[{\includegraphics[width=1in,height=1.25in,clip,keepaspectratio]{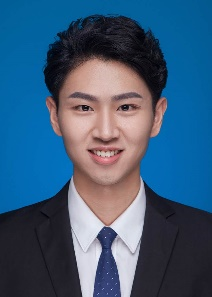}}]{Zhiyu Huang} received his B.E. degree from the School of Automobile Engineering, Chongqing University, Chongqing, China, in 2019. He is currently pursuing his Ph.D. degree with the School of Mechanical and Aerospace Engineering, Nanyang Technological University, Singapore. His current research interests include machine learning for decision-making and control and human-machine system with applications in automated driving.
\end{IEEEbiography}

\begin{IEEEbiography}[{\includegraphics[width=1in,height=1.25in,clip,keepaspectratio]{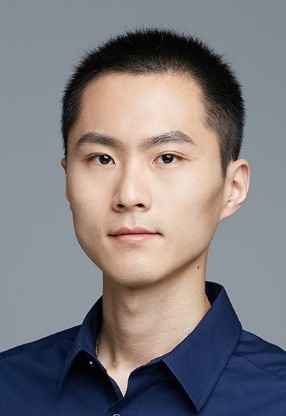}}]{Chen Lv} (M’16-SM’20) is currently an Assistant Professor at Nanyang Technology University, Singapore. He received the Ph.D. degree at the Department of Automotive Engineering, Tsinghua University, China in 2016. He was a joint Ph.D. researcher at EECS Dept., University of California, Berkeley, USA, during 2014-2015, and a Research Fellow with the Advanced Vehicle Engineering Center, Cranfield, University, U.K., during 2016 and 2018. His research focuses on advanced vehicles and human-machine systems, where he has contributed over 100 papers and obtained 12 granted patents. Dr. Lv serves as a Guest Editor for IEEE/ASME TMECH, IEEE ITS Magazine, Applied Energy, etc., and an Associate Editor/Editorial Board Member for International Journal of Vehicle Autonomous Systems, Frontiers in Mechanical Engineering, Vehicles, etc. He received the Highly Commended Paper Award of IMechE UK in 2012, the NSK Outstanding Mechanical Engineering Paper Award in 2014, the CSAE Outstanding Paper Award in 2015, Tsinghua University Outstanding Doctoral Thesis Award in 2016, the CSAE Outstanding Doctoral Thesis Award, and IEEE IV Best Workshop/Special Session Paper Award in 2018.
\end{IEEEbiography}

\begin{IEEEbiography}[{\includegraphics[width=1in,height=1.25in,clip,keepaspectratio]{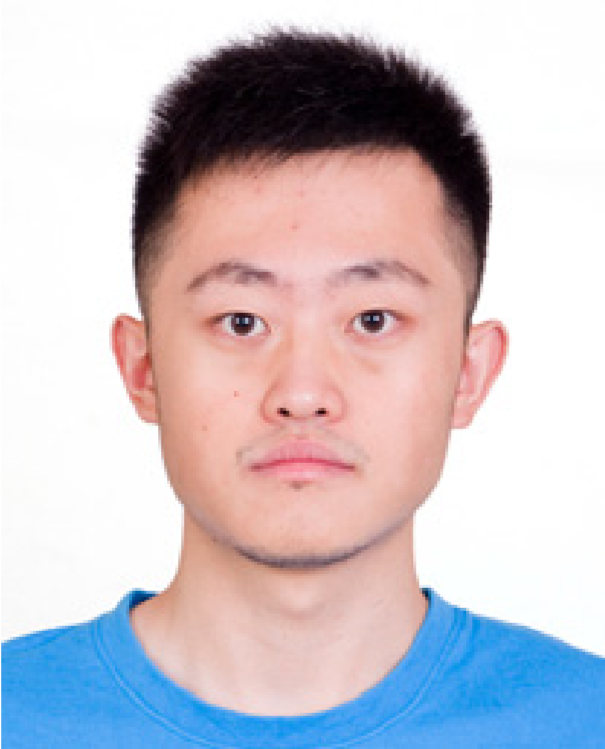}}]{Yang Xing} received his Ph.D. degree from Cranfield University, UK, in 2018. He is currently a Research Fellow with the Department of Mechanical and Aerospace Engineering at Nanyang Technological University, Singapore.  His research interests include machine learning, driver behavior modeling, intelligent multi-agent collaboration, and intelligent/autonomous vehicles. His work focuses on the understanding of driver behaviors using machine-learning methods and intelligent and automated vehicle design. He received the IV2018 Best Workshop/Special Issue Paper Award. Dr. Xing serves as a Guest Editor for IEEE Internet of Thing and Frontiers in Mechanical Engineering, and he is an active reviewer for IEEE Transactions on Intelligent Transportation Systems, Vehicular Technology, Industrial Electronics, and IEEE/ASME Transactions on Mechatronics.
\end{IEEEbiography}

\begin{IEEEbiography}[{\includegraphics[width=1in,height=1.25in,clip,keepaspectratio]{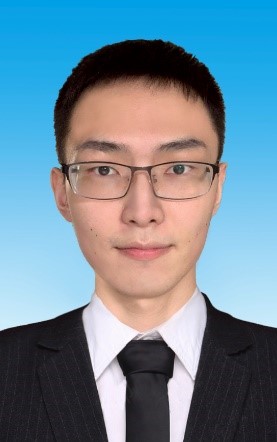}}]{Jingda Wu} received his B.S. (2016) and M.S. (2019) in mechanical engineering from Beijing Institute of Technology, China. He is currently working on his Ph.D. in mechanical engineering at Nanyang Technological Univ., Singapore. His research mainly focuses on control and optimization of human machine collaborated driving, machine learning techniques, design of autonomous driving strategy, energy management of electric vehicle and Li-ion battery.
\end{IEEEbiography}

\end{document}